# Multi-visual modality micro drone-based structural damage detection

Isaac Osei Agyemang[a,*], Liaoyuan Zeng[b,c], Jianwen Chen[a], Isaac Adjei-Mensah[a], Daniel Acheampong[d]

[a] *School of Information and Communication Engineering, University of Electronic Science and Technology of China, Chengdu 610054, China*
[b] *Yibin Institute of University of Electronic Science and Technology of China*
[c] *Intelligent Terminal Key Laboratory of Sichuan Province*
[d] *Lutgert College, Florida Gulf Coast University, Florida, USA*

**Abstract:** Accurate detection and resilience of object detectors in structural damage detection are important in ensuring the continuous use of civil infrastructure. However, achieving robustness in object detectors remains a persistent challenge, impacting their ability to generalize effectively. This study proposes DetectorX, a robust framework for structural damage detection coupled with a micro drone. DetectorX addresses the challenges of object detector robustness by incorporating two innovative modules: a stem block and a spiral pooling technique. The stem block introduces a dynamic visual modality by leveraging the outputs of two Deep Convolutional Neural Network (DCNN) models. The framework employs the proposed event-based reward reinforcement learning to constrain the actions of a parent and child DCNN model leading to a reward. This results in the induction of two dynamic visual modalities alongside the Red, Green, and Blue (RGB) data. This enhancement significantly augments DetectorX's perception and adaptability in diverse environmental situations. Further, a spiral pooling technique, an online image augmentation method, strengthens the framework by increasing feature representations by concatenating spiraled and average/max pooled features. In three extensive experiments: (1) comparative and (2) robustness, which use the Pacific Earthquake Engineering Research Hub ImageNet dataset, and (3) field-experiment, DetectorX performed satisfactorily across varying metrics, including precision (0.88), recall (0.84), average precision (0.91), mean average precision (0.76), and mean average recall (0.73), compared to the competing detectors including You Only Look Once X-medium (YOLOX-m) and others. The study's findings indicate that DetectorX can provide satisfactory results and demonstrate resilience in challenging environments.

## 1. Introduction

The continuous operation and safety of civil infrastructures, which include essential structures like buildings, bridges, dams, and tunnels, are of utmost importance for the efficient functioning of society. However, these infrastructures are consistently subjected to various environmental stresses, encompassing natural disasters such as earthquakes and floods and the natural deterioration resulting from material degradation over an extended period [1]-[2]. Further, continuous exposure to environmental stresses can lead to structural deficiencies and irregularities, compromising these essential civil infrastructures' overall strength and durability. Also, the repercussions of the compromised structural health of these civil infrastructures potentially have disastrous effects on human lives, the environment, and the economy [3]. Implementing robust structural health monitoring (SHM) systems is essential in addressing these potential hazards and maintaining the operational integrity of civil infrastructures. Previously, the usual approaches for evaluating the health of infrastructure after construction utilized physical inspections and traditional image processing techniques, which have demonstrated inefficiency, imprecision, and high resource demand in most cases [4]–[7]. Moreover, the intricate nature and vast magnitude of specific infrastructures, such as skyscrapers and expansive bridge structures, provide considerable obstacles for precise and prompt human-visual inspection.

Recently, the emergence of Deep Convolutional Neural Networks (DCNN) in computer vision (CV) tasks such as classification, detection, and segmentation has brought about a paradigm shift in the field of structural health monitoring (SHM) as these artificial intelligence techniques offer an efficient assessment of varying structural damages [8]–[13]. Utilizing deep learning and computer vision enables DCNNs to potentially overcome the constraints of conventional techniques and positively impact structural health monitoring. Regardless of the promising performance of DCNN in SHM variant tasks, there are occasional sub-optimal performances, arguably attributed to the less robustness of the DCNN model due to image environments, such as shadows, reflections, occlusions, illumination, and other visual elements, which potentially limits the percepts of DCNN. As a result, varying image augmentation techniques have been introduced into the training of DCNN models to help improve their robustness; however, most of these augmentation methods are preprocesses and not online during a model's training [14]–[16]. Irrespective of the marginal to moderate improvement image augmentation offers, the less robustness of DCNN models in certain CV tasks, such as object detection in SHM, persists. Further, integrating attention mechanisms, which significantly boost the performances of DCNN, is also viable; however, extreme cases of poor image quality, such as heavy blurring or extreme noise, could affect the attention mechanism's overall performance [17]-[18]. Consequently, if the image quality is so severe that important features become indistinguishable or distorted beyond recognition, the attention mechanism's ability to allocate weights properly could be compromised.

*Corresponding author: School of Information and Communication Engineering, University of Electronic Science and Technology of China.
Email address: 45ioa@uestc.edu.cn (I.O. Agyemang)

There has been a surge in the integration of DCNN and varying smart devices, including contact (e.g., displacement sensors, load cells, strain gauges) and non-contact (e.g., vibration sensors, unmanned aerial vehicles) devices, to enhance the automation aspect of SHM further. In SHM, considerable interest is in unmanned aerial vehicles (UAVs), primarily with much emphasis on the larger UAVs (macro UAVs), neglecting the potential of the smaller UAVs (nano-to-micro UAVs) [19]. Macro UAVs possess high precision and adaptability and integrate well with varying sensor modalities; hence, they are well-suited for inspecting extensive or sizable structures that necessitate thorough and efficient inspections [20]–[23]. Nevertheless, it is imperative to acknowledge that Macro UAVs offer unique benefits and considerations applicable to SHM tasks. Micro UAVs can navigate environments with narrow passages, gain entry into hard-to-reach regions (e.g., space between smaller piers), and exhibit enhanced agility during operation [24]–[26]. Additionally, agility is highly helpful in inspecting structural elements, including internal spaces, complex geometry, and structures in crowded surroundings [27]. These potentials of micro UAVs provide distinct benefits that can supplement the capabilities of macro UAVs. Regardless of the potential of micro UAVs, there has been little integration in SHM tasks.

To this end, this study proposes a framework, DetectorX, to improve the percept and robustness capabilities of DCNN models based on the proposition of event-based reward reinforcement learning and spiral pooling. Additionally, DetectorX facilitates the use of micro UAVs for automated SHM, specifically damage detection post-construction. The main contributions of the study are as follows:

1. A vision-based adaptive framework, DetectorX, which comprises two DCNN models, is proposed to increase the perception. One DCNN model, leveraging a secondary proposition, event-based reward reinforcement learning, dynamically induces a visual modality to enhance the framework's perceptual capabilities. The second DCNN model functions as an object detector. This module directly aids the framework in dynamically learning different visual modalities, thereby improving its perception and versatility in handling diverse environmental scenarios.
2. A generic and adaptive feature map augmentation, spiraling pooling, which induces additional robustness and localized context, is proposed, which can be seamlessly integrated into other DCNN-based models. This contribution directly tackles the challenge of offline image augmentation methods by having an online augmentation method and a versatile augmentation approach that improves the overall robustness of object detection models.
3. Lastly, the validation of the proposed framework, DetectorX, is assessed by utilizing a micro UAV to provide empirical evidence of its practicality and viability for autonomous inspection and evaluation of civil infrastructure. This real-world validation contributes to bridging the gap between theoretical frameworks and practical implementation, demonstrating the applicability and potential impact of DetectorX in real-world scenarios.

The remaining content of the paper is as follows: Sections 2 and 3 detail the literature review and proposed DetectorX framework, respectively. Section 4 presents the study's experiment, while Section 5 discusses the results. Lastly, Section 6 concludes the study.

## 2. Literature review

### *2.1. DCNN-based detection methods for SHM*

Structural health monitoring is one field that has seen continuous growth in the application of DCNNs revolving around computer vision tasks such as object detection, classification, recognition, and segmentation [9]-[10]. Among these tasks, object detection is vital in SHM due to the localization and classification of the object of interest, such as structural damages.

Object detection has been crucial in SHM regarding varying civil infrastructures [28]–[33]. Recently, a study presented an improved YOLOv7 architecture incorporating three innovative self-designed modules to mitigate the challenge of detecting concrete cracks [34]. The enhanced YOLOv7, Neue Modultechnologie, integrated the SwinTransformer and utilized residual connections to address perceptual scope, accuracy, and gradient optimization challenges. The proposed network demonstrated promising detection results on a custom crack dataset; however, the custom dataset used was clean. Dang et al. proposed a new transformer-based model for efficient concrete damage detection [35]. The proposed framework is based on Deformable DETR, incorporating additional modules into Deformable DETR to enhance its performance. The proposed architecture demonstrates promising performance on the concrete crack dataset compared to some object detection models, obtaining a mean Average Precision (mAP) of 63.8%. The study acknowledges the limitation of the inability to detect real-time defects and the proposed network's lack of robustness. Also, crack detection with limited data was studied in [36]. The proposed network involved the application of transfer learning, leveraging a minimum amount of annotated data. Subsequently, the model was fine-tuned and optimized to focus on detecting cracks using the brickwork crack dataset. The resulting section of the study demonstrates the promising performance of the authors' proposition; however, the study did not show any real-world application.

Further, an improved YOLOv5 model with an attention mechanism was used to detect salient concrete surface defects utilizing the SDNET 2018 concrete crack dataset [37]. The resulting section of the study demonstrates the promising performance of the improved YOLOv5 model; however, extensive experiment regarding robustness to demonstrate the model's performance against salient cracks was not given. Also, a hybrid deep learning network that combines the convolutional neural network and Long Short Term Memory (LSTM) network was proposed to detect structural damages [38]. The hybrid network was implemented to assess the structural deterioration of three distinct self-compacting concrete samples. The experiment and analysis using a custom dataset demonstrate that the proposed network performs better than conventional deep-learning approaches. The limitation associated with the study is the reduction in inference time due to the fusion of the LSTM and the CNN model. Weng et al. introduced DACrack, an unsupervised domain adaptation framework for civil infrastructure crack detection [39]. DACrack employs a combination of contrastive processes, adversarial learning, and variational autoencoders to effectively execute domain adaptation across the input, feature, and output domains. Extensive experiments on four datasets, namely G45, CFD, CrackFCN, and CrackTunnel, exhibited the promising performance of DACrack. Although DACrack attained promising performances, DACrack relies on careful selection of source domains and multiple training to yield the desired results. Ye et al. presented a structural damage detection method utilizing a fully convolutional neural network [15]. The study utilized a custom dataset of interior concrete cracks and attained a detection accuracy of 93.6%. The conclusion section of the study admits the proposed framework localization of objects of interest fails if there are no geo-distinctive visual elements in an image. Likewise, a supervised deep convolutional neural network was employed by fusing two varying models to detect various forms of damage in challenging structural scenarios where custom UAV data was utilized [40]. The accuracy of damage detection reported in the study was 80.7%; however, the proposition requires extra preprocessing to suit the model's specifications prior to inference.

Although most of the recent DCNN-based detectors in SHM tasks attain promising performances, much attention has not been paid to robustness and image environments (e.g., shadows, illumination, reflections, and other visual elements), as evident in the reiterated studies. These image environments can be used as augmentation mechanisms to boost the robustness of DCNN models and, at the same time, can also undermine their performances.

### *2.2. DCNN-based detection methods with UAVs for SHM*

Unmanned aerial vehicles (UAVs) continuously dominate structural health monitoring due to their aerial advantage, large terrain coverage, and other essential attributes.

Recently, an inference model based on the U-Net architecture was employed to analyze a DJI Matrice (UAV) video stream to detect structural defects in pavements [41]. The proposed model was trained and tested with custom data. Although satisfactory performance is reported in the study, the study acknowledges slow inference of the proposed model, hence not making the proposed model ideal in real-time applications. Similarly, A DJI Matrice RTK 300 equipped with a Zenmuse H20T laser range finder was used in a UAV-based detection and quantification of concrete cracks [42]. The study also introduces an independent boundary refinement transformer to improve the detection of cracks. The study utilized a custom dataset, which claimed complex image environments were present, yet no robustness evaluation was conducted. In another study, a modified version of the Faster-RCNN was used as an inference model with a UAV integrated with Microsoft Kinect V2 RGB-D camera to detect and classify structural damages [43]. The data collected via the Microsoft Kinect V2 RGB-D camera, which adds depth to the data, served as the training and testing data. Although the resulting section of the study is interesting and promising, further evaluation is required to validate the proposition further. Two fabricated UAVs equipped with a stereo camera, lidar, and an NVIDIA Jetson Xavier NX were used in a UAV-based explore-then-exploit vision-based task [22]. The proposition incorporates a two-step simultaneous localization and mapping (SLAM) technique for determining the position of the UAVs and creating a three-dimensional representation of the environment. The data from the ground station goes through an augmentation process, where varying temperatures are added to the data, serving as training and testing data for the proposed encoder-decoder framework, which was used for inference of damage detection. Additionally, it includes a coverage path planning algorithm that ensures safety during inspection and data collection activities and a trajectory generation method that considers the presence of obstacles. The proposed framwork process of data capture to inference making is time consuming due to the temperature augmentation step. Additionally, the proposition is only tailored for enconder-decoder models and cannot be extended to other DCNN-based model architectures. In a similar study, an Unmanned Aerial System (UAS) equipped with five primary components, namely a main frame, motors, a flight controller, a camera, and a data transmitter, was used with a CNN model to detect cracks on concrete surfaces [44]. The CNN model operated on an Android device, which received the streaming data from the UAS. The inference model was first pre-trained on the crack dataset before additional custom data training. Although the study demonstrated detailed practicability, the setup for inspection requires more than one personnel, and it is also costly. Also, a drone with a DCNN model as an inference mechanism was used to autonomously navigate and inspect reinforced concrete structures indoors [45]. The automation aspect of the study incorporated obstacle avoidance, which was facilitated by geotagging. The inference model was pre-trained on non-structural damage-related images before being trained on a custom structural damage dataset. The results reported in the study show varying promising levels; however, the proposed pipeline for damage detection is associated with a limited detection range. Also, a manually controlled micro UAV was utilized to inspect and assess multi-attribute structural damage detection [46]. The authors' proposition comprised three DCNN models, one serving as a classifier and the other as an ensemble detector. The study utilized the PEER Hub ImageNet multi-attribute dataset to evaluate and analyse the proposed EnsembleDetNet framework. Although the proposed framework attained desirable results, the computational load of the framework needs further optimization.

As evident in the above literature, macro UAVs have dominated SHM; however, micro UAVs can be used equally in most SHM tasks due to their agility, ability to navigate narrow terrains, and other favorable attributes. Moreover, micro UAVs are less costly and easily operated by a small team of engineering practitioners, which is beneficial in SHM as it mitigates the cost of conducting structural health monitoring, inspection, and assessment.

## 3. Proposed DetectorX framework

### 3.1 *DetectorX framework* overview

The proposed framework, DetectorX, has two adapted DCNN models, EfficientNet-B4 [47] and EfficientDet-D4 [48]. These two models were selected due to their efficient nature regarding computation. EfficientNet-B4 is implemented based on the proposed event-based reward reinforcement learning, which constrains parent and child actions to aid in inducing dynamic visual modalities via a stem block to DetectorX during training, unlike other visual enhancements and augmentation techniques, classified as preprocessing techniques. Also, the proposed spiral pooling mechanism, which induces varied feature maps, is introduced into the two DCNN models in the DetectorX framework to increase robustness.

Specifically, the modified EfficientNet-B4 has a dual fully connected layer with a regressor, which predicts two tuples $(T_e, D_t)$. Tuple $T_e = (colormap_{value}, alpha_{value})$ entails numerical predictions for mapping selected OpenCV colormap types and an alpha value for adjusting transparency to mimic a thermal-like effect on the input data. Similarly, tuple $D_t = (brightness_{value}, contrast_{value}, gamma_{value}, saturation_{value}, hueshift_{value})$ are the numerical predictions, which create a dark theme overlay on the input data to induce an additional visual modality. By altering the input data based on the thermal-like effects and the dark theme, two varied visual modalities are induced via a stem block as a concatenated feature map in addition to the RGB visual modality being fed to the adapted EfficientDet-D4 as visualized in Fig. 1.

Finally, the adapted EfficientDet-D4 detects structural damages by learning from three visual modalities: the RGB, thermal-like effect, and dark theme modalities. These three visual modalities and the spiral pooling mechanism improve the percept and robustness of the proposed DetectorX framework, making it adapt to varying scenes. Fig. 1 illustrates the proposed DetectorX framework, and Algorithm 1 delineates the abstraction level of DetectorX for inference. Fig. 2 shows random samples of input data and their varied visual modalities and the average pooled feature map. Table 1 summarizes the prediction range of the adapted EfficientNet-B4. Next, the proposed stem block and spiral pooling is given in sub-sections 3.2 and 3.3, respectively; this is followed by the adapted DCNN models in DetectorX in sub-section 3.4. Afterward, the proposed event-based reward reinforcement learning, which DetectorX operates on, is given in sub-section 3.5.

**Table 1**
The prediction range and mapping of the modified EfficientNet-B4.

| Tuple $T_e$ | Prediction mapping/range value | Tuple $D_t$ | Prediction range value |
|---|---|---|---|
| $colormap_{value}$ = JET | 1 | $brightness_{value}$ | 0.0 to1.0 |
| $colormap_{value}$ = RAINBOW | 2 | $contrast_{value}$ | 0.0 to1.0 |
| $colormap_{value}$ = HSV | 3 | $gamma_{value}$ | 0.0 to1.0 |
| $colormap_{value}$ = TURBO | 4 | $saturation_{value}$ | 0.0 to1.0 |
| $alpha_{value}$ | 0.0 to1.0 | $hueshift_{value}$ | 0.0 to1.0 |

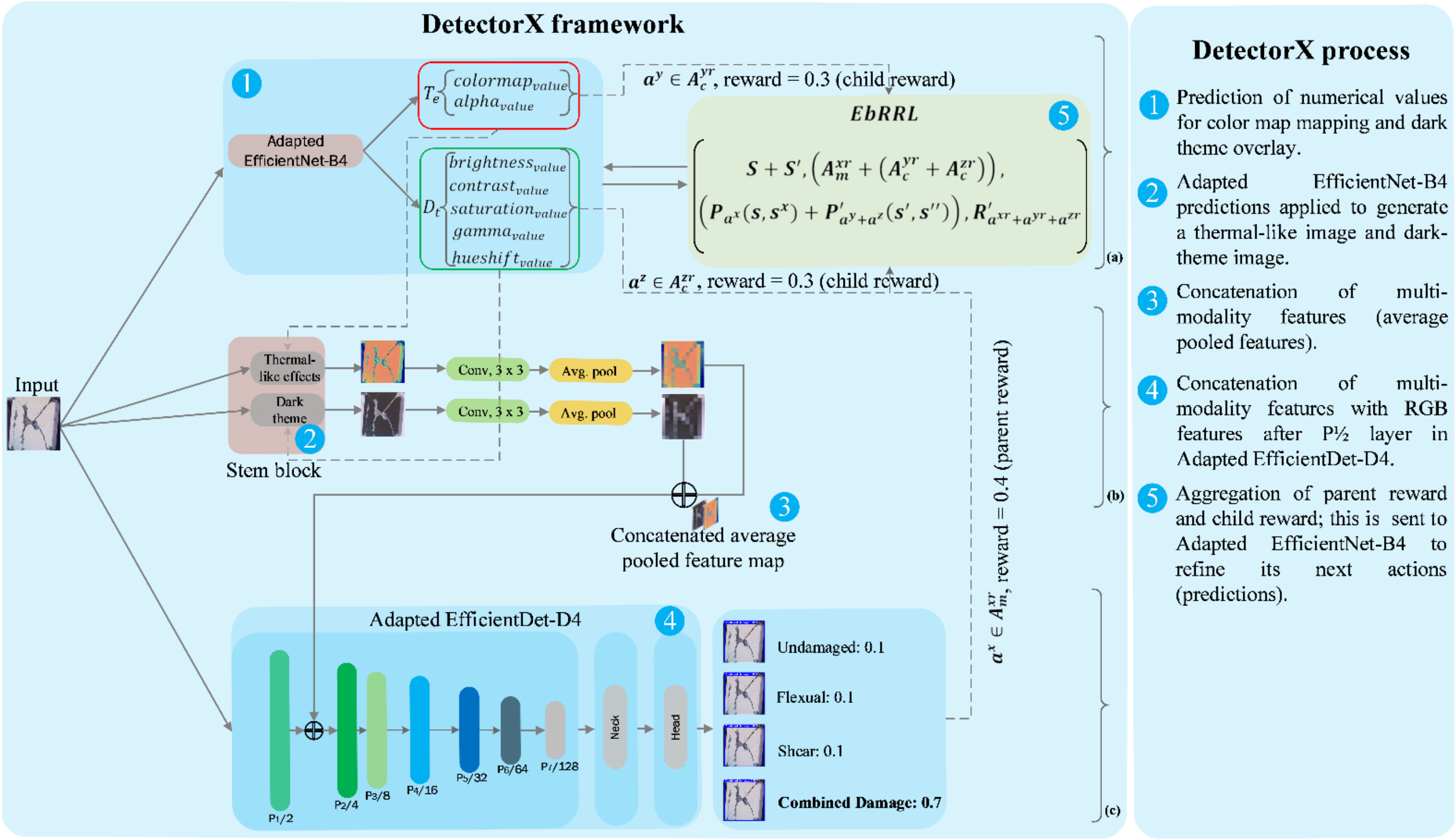

**Fig. 1.** Overview of the proposed DetectorX framework: (a) adapted EfficientNet-B4 implemented based on the proposed event-based reward reinforcement learning, (b) the stem block which induces thermal-like and dark theme visual modalities based on the predictions of the adapted EfficientNet-B4, and (c) adapted EfficientDet-D4 for structural damage detection.

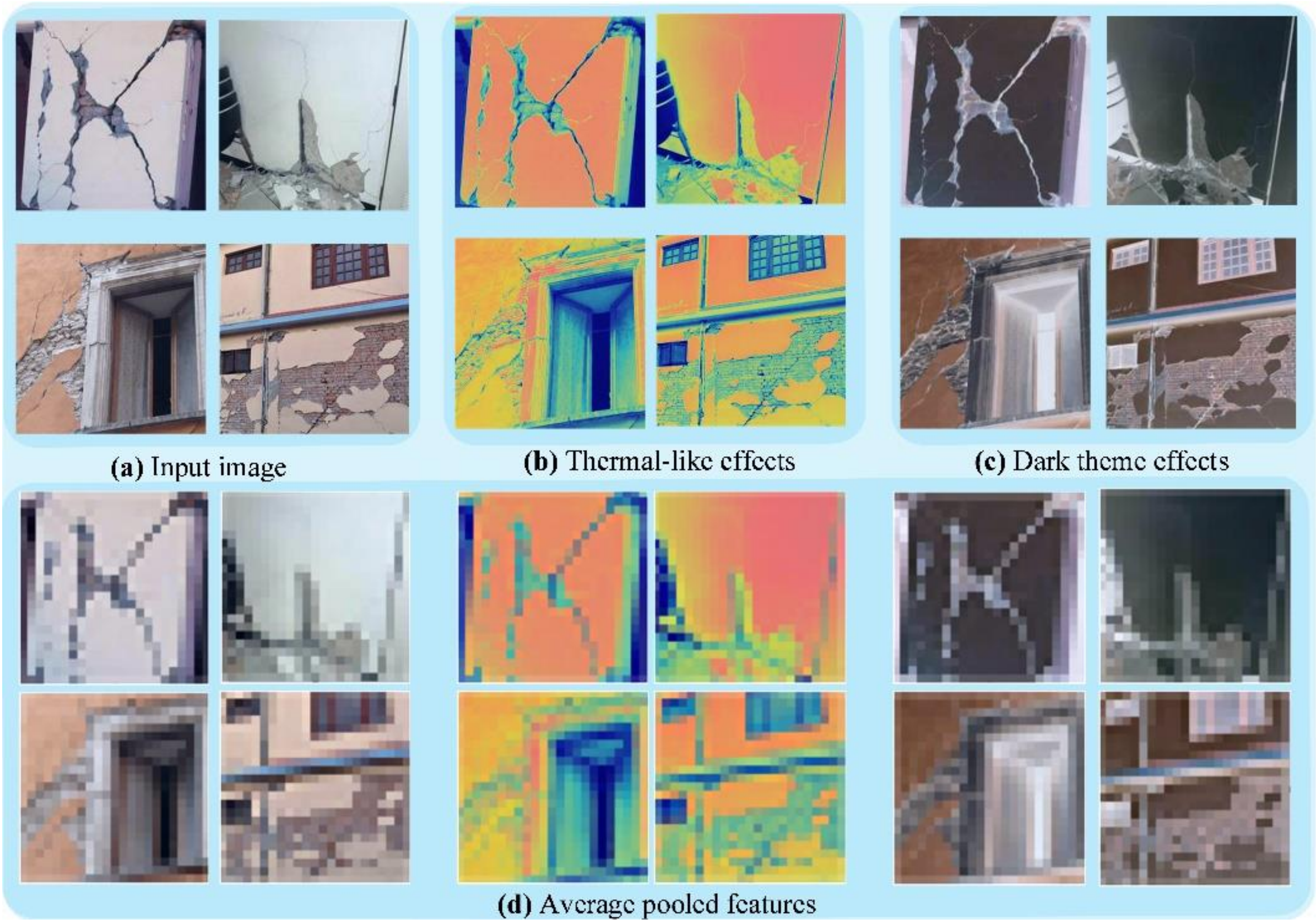

**Fig. 2.** Random samples of input data and their corresponding visual modality based on the adapted EfficientNet-B4 predictions and the average pooled features from the stem block.

**Algorithm 1:** DetectorX for structural damage detection

***# Initialize DetectorX***

***# Connect to video/image source***

**While** (input_img != null)

DetectorX ← input_img

***# Forward pass through EfficientNet-B4***

EfficientNet-B4 in DetectorX→ $(T_e, D_t)$

$T_e = (colormap_{value}, alpha_{value})$

$D_t = (brightness_{value}, contrast_{value}, gamma_{value}, saturation_{value}, hueshift_{value})$

***# Stem-block adjust input_img based on*** $(\boldsymbol{T_e}, \boldsymbol{D_t})$

Stem-block → thermal-like features $TL_{map}$ and dark theme features $DT_{map}$ (multi-modality features)

***# Concatenate multi-modality features with RGB features*** $\boldsymbol{RGB_{map}}$ ***in EfficientDet-D4 at layer*** $\boldsymbol{P_{1/2}}$.

Concatenated feature maps = $TL_{map} \oplus DT_{map} \oplus RGB_{map}$

EfficientDet-D4 in DetectorX → structural damage prediction

***#Aggregation of parent and child rewards***

$EbRRL = \left(S + S', \left(A_m^{xr} + \left(A_c^{yr} + A_c^{zr}\right)\right), \left(P_{a^x}(s, s^x) + P'_{a^y + a^z}(s', s'')\right), R'_{a^{xr} + a^{yr} + a^{zr}}\right)$

***# Update EfficientNet-B4 in DetectorX with a reward from EbRRL.***

EfficientNet-B4 in DetectorX ← $R'_{a^{xr} + a^{yr} + a^{zr}}$ (reward)

**Else**

Terminate DetectorX

**End**

### *3.2 Stem block in DetectorX*

The stem block in DetectorX plays a pivotal role in dynamically adjusting the input image based on the predictions of the adapted EfficientNet-B4 model in DetectorX. This adaptive online augmentation processing step induces two distinct visual modalities—thermal-like effect and dark theme overlay—enhancing the framework's perceptual capabilities. The stem block consists of several processes.

First, the stem block feeds on the outputs of the adapted EfficientNet-B4, which produces two sets of numerical predictions, denoted as tuples $T_e = (colormap_{value}, alpha_{value})$ and $D_t = (brightness_{value}, contrast_{value}, gamma_{value}, saturation_{value}, hueshift_{value})$, respectively. Tuple $T_e$ entails information necessary for generating a colormap-based thermal-like effect and tuple $D_t$ predictions contain parameters for creating a dark theme overlay on the input image. The colormap value from $T_e$ is utilized to map selected OpenCV colormap types in Table 1 onto the input image, creating a visual representation reminiscent of thermal imaging. The alpha value in $T_e$ adjusts transparency, effectively mimicking the thermal-like effect. Tuple $D_t$ parameters contribute to the generation of a dark theme overlay on the input image. Brightness, contrast, gamma, saturation, and hue adjustments collectively create a visually distinct dark-themed representation. The stem block processes the input image based on the generated thermal-like effect feature map $TL_{map}$ and dark theme overlay feature map $DT_{map}$. The resulting multi-modality features ($TL_{map}$ and $DT_{map}$), representing thermal-like and dark theme modalities/features, are then concatenated and fed into the subsequent layers of the adapted EfficientDet-D4 in DetectorX, where it is feature concatenated with the RGB feature map $RGB_{map}$ after layer $P_{1/2}$ in the adapted EfficientDet-D4, as seen in Fig. 1.

The adaptive nature of the stem block, guided by the adapted EfficientNet-B4's predictions, allows DetectorX to learn and adapt dynamically to various environmental scenarios. This multi-modal approach significantly enhances the model's robustness and perceptual capabilities, contributing to its effectiveness in structural damage detection.

### *3.3 Spiral pooling*

The robustness, improved performance, and generalization capabilities of DCNN models can be attributed to the use of augmented training data. The predominant image augmentation methods can be classified as preprocessing since they precede the training stage of DCNN models. These conventional augmentation techniques are typically performed before training, making them an offline approach, constraining the initial orientation of the features within the training data; hence, the training data features maintain consistent orientation and position across the layers or blocks of convolutions. This study hypothesized that the performance of DCNN models could be further enhanced by having an online feature augmentation, thus augmentation during the training of a DCNN model. As a result, spiral pooling, which stems from a spiral matrix, is proposed as an online image augmentation mechanism. Spiral pooling aims to induce robustness by performing spiral matrix operations on feature maps. This technique capitalizes on spiral patterns' inherent structure and characteristics, which inherently encapsulate progressive variation and multi-directionality. Spiral matrix operations can increase an image's intricate feature orientation, contextual information, and spatial relationships.

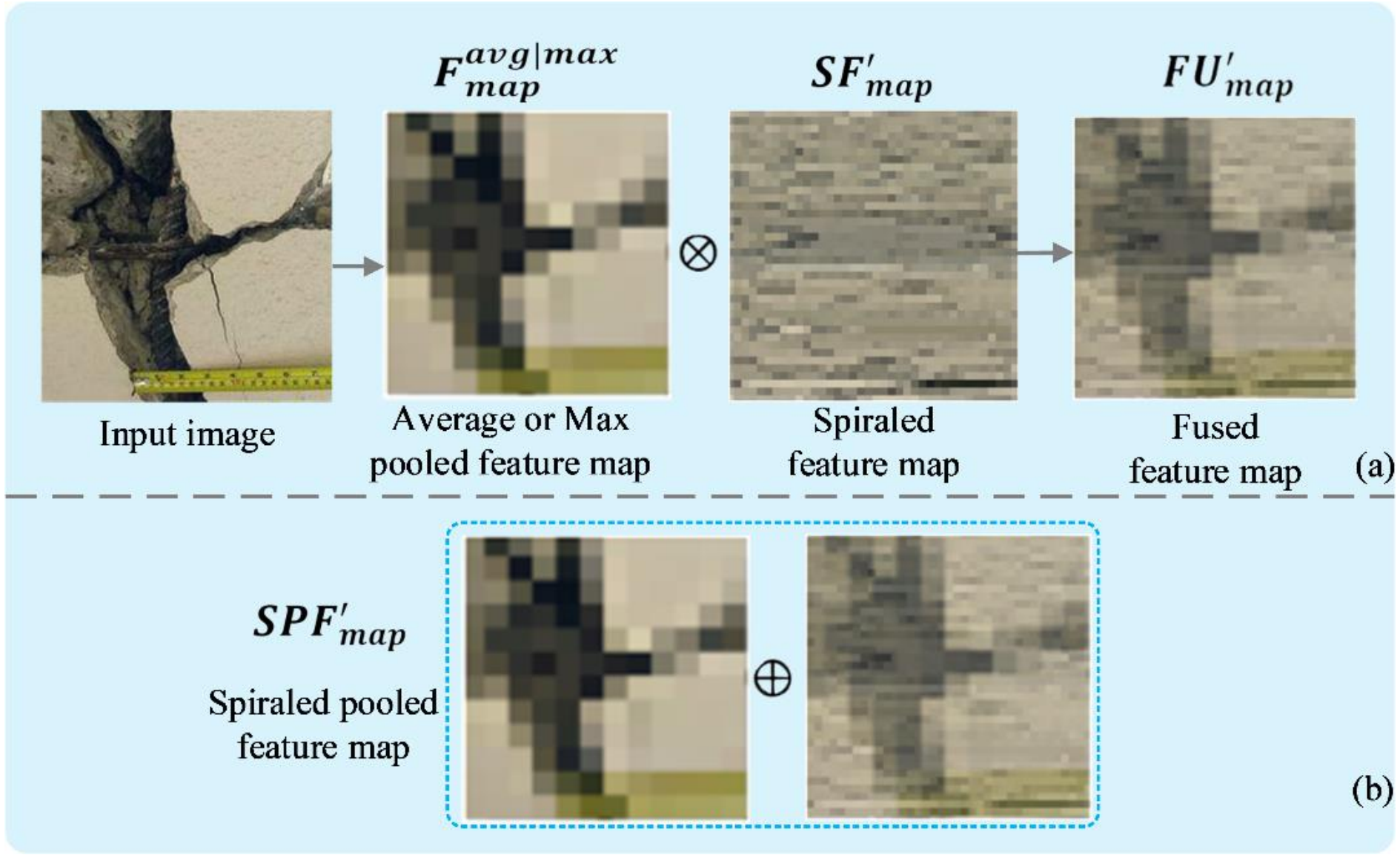

Fig. 3. Graphical abstraction of the process of the proposed spiral pooling: (a) element-wise fusing of spiraled feature map with average or max pooled feature map, and (b) concatenation of fused feature map with average or max pooled feature map.

The proposed spiral pooling works as follows:

1. A spiral matrix operation is performed on a maximum or average pooled feature map $\boldsymbol{F^{avg|max}_{map}}$ to adduce spiraled feature map $\boldsymbol{SF'_{map}}$.
2. Afterwards, the spiraled feature map $\boldsymbol{SF'_{map}}$ is fused with the previous maximum or average pooled feature map $\boldsymbol{F^{avg|max}_{map}}$ via an element-wise operation.
3. Lastly, the fused feature map $\boldsymbol{FU'_{map}}$ is concatenated with the previous maximum or average pooled feature map $\boldsymbol{F^{avg|max}_{map}}$ before propagating to the next convolution layer or block.

As a practical example, given a maximum or average pooled feature map $\boldsymbol{F^{avg|max}_{map}}$**,** as Eq. (1):

$$\boldsymbol{F^{avg|max}_{map}} = \begin{bmatrix} \boldsymbol{P^{r1}_1}, & \boldsymbol{P^{r1}_2}, \cdots, & \boldsymbol{P^{r1}_n} \\ \boldsymbol{P^{r2}_1}, & \boldsymbol{P^{r2}_2}, \cdots, & \boldsymbol{P^{r2}_n} \\ \vdots & \vdots \quad \ddots & \vdots \\ \boldsymbol{P^{rn}_1}, & \boldsymbol{P^{rn}_2}, \cdots, & \boldsymbol{P^{rn}_n} \end{bmatrix} \tag{1}$$

where $\boldsymbol{P^{rn}_n}$ denotes the indices of the pixel of the feature map. The spiraled feature map $\boldsymbol{SF'_{map}}$ can be adduced by performing a spiral operation with a time complexity of $O(m \times n)$ and space complexity of $O(n)$ on $\boldsymbol{F^{avg|max}_{map}}$ as follows:

1. Traverse the first row of Eq. (1) from left to right ( $\boldsymbol{P^{r1}_1}$ to $\boldsymbol{P^{r1}_n}$).
2. Traverse the last column of Eq. (1) from top to bottom ( $\boldsymbol{P^{r1}_n}$ – 1 index to $\boldsymbol{P^{rn}_n}$).
3. Traverse the last row of Eq. (1) from right to left ( $\boldsymbol{P^{rn}_n}$ – 1 index to $\boldsymbol{P^{rn}_1}$).
4. Traverse the first column of Eq. (1) from bottom to top ( $\boldsymbol{P^{rn}_1}$ – 1 index to 1 index – $\boldsymbol{P^{r1}_1}$).

The above spiral operation performed on Eq. (1) results in the spiraled feature map $\boldsymbol{SF'_{map}}$**,** as in Eq.(2).

$$\boldsymbol{SF'_{map}} = \begin{bmatrix} \boldsymbol{P^{r1}_1}, \boldsymbol{P^{r1}_2}, \cdots, \boldsymbol{P^{r1}_n} \\ \boldsymbol{P^{r2}_n}, \quad \vdots \ , \boldsymbol{P^{rn}_n}, \cdots \\ \boldsymbol{P^{rn}_2}, \ \boldsymbol{P^{rn}_1}, \ \vdots \ , \boldsymbol{P^{r2}_1} \\ \boldsymbol{P^{r2}_2}, \cdots, \quad \ddots \ , \ \vdots \end{bmatrix} \tag{2}$$

Afterwards, the spiraled feature map $\boldsymbol{SF'_{map}}$ is fused with the average or max pooled feature map $\boldsymbol{F^{avg|max}_{map}}$ to attain a fused feature map $\boldsymbol{FU'_{map}}$**,** as in Eq. (3).

$$\boldsymbol{FU'_{map}} = \begin{bmatrix} \boldsymbol{P_1^{r1}},\ \boldsymbol{P_2^{r1}}, \cdots,\ \boldsymbol{P_n^{r1}} \\ \boldsymbol{P_n^{r2}},\ \ \vdots\ ,\ \boldsymbol{P_n^{rn}}, \cdots \\ \boldsymbol{P_2^{rn}},\ \boldsymbol{P_1^{rn}},\ \vdots\ ,\boldsymbol{P_1^{r2}} \\ \boldsymbol{P_2^{r2}}, \cdots,\ \ \ddots\ \ ,\ \vdots \end{bmatrix} \otimes \begin{bmatrix} \boldsymbol{P_1^{r1}},\ \boldsymbol{P_2^{r1}}, \cdots,\ \boldsymbol{P_n^{r1}} \\ \boldsymbol{P_1^{r2}},\ \boldsymbol{P_2^{r2}}, \cdots,\ \boldsymbol{P_n^{r2}} \\ \vdots \quad \vdots \quad \ddots \quad \vdots \\ \boldsymbol{P_1^{rn}},\ \boldsymbol{P_2^{rn}}, \cdots,\ \boldsymbol{P_n^{rn}} \end{bmatrix} \tag{3}$$

Lastly, to deduce the proposed spiral pooling feature map $\boldsymbol{SPF'_{map}}$, the fused feature map $\boldsymbol{FU'_{map}}$ is concatenated with the previous maximum or average pooled feature map $\boldsymbol{F_{map}^{avg|max}}$, as in Eq. (4).

$$\boldsymbol{SPF'_{map}} = \boldsymbol{FU'_{map}} \oplus \boldsymbol{F_{map}^{avg|max}} \tag{4}$$

Spiral Pooling can be integrated seamlessly into varying DCNN architecture as a supplementary layer following convolutional operations to extract more comprehensive feature representations encompassing a wider range of orientations and contextual cues. This diverse set of features contributes to a more robust understanding of complex visual environments. Fig. 2 gives a graphical insight into the procedure of the proposed spiral pooling.

### *3.4 Adapted DCNN models in DetectorX*

EfficientNet-B4 and EfficientDet-D4 are utilized as the DCNN models in DetectorX, with spiral pooling plugged in both DCNN models in addition to the dynamic visual modalities for robustness.

The EfficientNet series is a family of neural network architectures designed to achieve state-of-the-art performance while being computationally efficient. The key innovation behind EfficientNet is using a compound scaling method that systematically balances model depth, width, and resolution; this allows EfficientNet models to achieve better accuracy than some traditional DCNN models while using fewer parameters and less computation. EfficientNet introduces three scaling dimensions:

1. Depth scaling: Increases the number of layers in the network to aid the network in capturing more complex features.
2. Width scaling: Increases the width of the network by expanding the number of channels in each layer; this allows the network to capture more fine-grained features.
3. Resolution scaling: Increases the input image resolution to provide more details to the network.

Also, EfficientNet uses a compound scaling formula that scales these dimensions (the model depth, width, and resolution, respectively) while keeping the model's computational cost relatively constant.

On the other hand, EfficientDet is an extension of the EfficientNet architecture specifically designed for object detection tasks. EfficientDet leverages the efficiency and effectiveness of EfficientNet and introduces modifications to adapt it to the object detection domain. The key features of EfficientDet are as follows:

1. Bidirectional Feature Pyramid Network (BiFPN): A novel feature fusion method that combines multi-level features and creates a feature pyramid; this helps the model capture objects of different scales and improve localization accuracy.
2. Compound scaling: Like EfficientNet, EfficientDet uses compound scaling to balance depth, width, and resolution; this ensures that the model can handle a wide range of object sizes.
3. Anchor box scales and ratios: EfficientDet employs different anchor box scales and ratios for each model variant, improving the model's ability to detect objects of various shapes and sizes.
4. Task-Specific Heads: EfficientDet includes task-specific heads for class prediction and bounding box regression. These heads are attached to different levels of the feature pyramid, allowing the model to learn high-level and low-level features for accurate object detection.

EfficientDet models, denoted as EfficientDet-D0, EfficientDet-D1, ..., and EfficientDet-D7, offer a trade-off between speed and accuracy. Smaller variants are faster but may compromise some precision, while larger variants are more accurate but require more computational resources.

In this study, EfficientNet-B4 and EfficientDet-D4 are used with the proposed spiral pooling in a vision-based task. The EfficientNet-B4 is used to predict numerical parameters to induce varying visual modalities in the DetectorX framework, while the EfficienDet-D4 takes advantage of these varying visual modalities in detecting objects of interest. Fig. 4 illustrates the adapted EfficientNet-B4 and EfficientDet-D4, respectively, with the proposed spiral pooling.

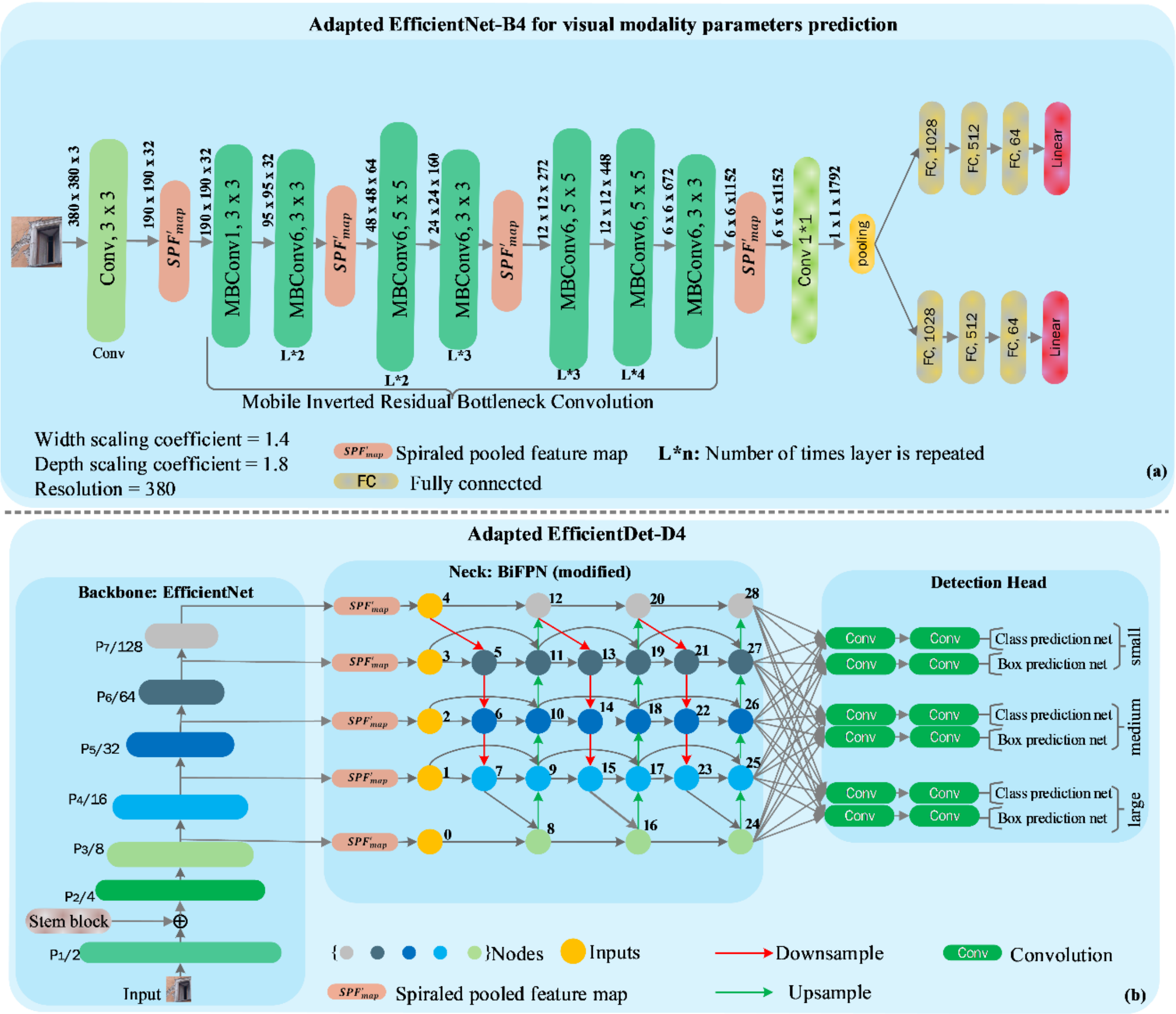

**Fig. 4.** Illustration of the adapted DCNN models in DetectorX framework: (a) adapted EfficientNet-B4, and (b) adapted EfficientDet-D4.

### 3.5 *Event-based reward reinforcement learning (EbRRL)*

The reinforcement learning training method for DCNN models, based on rewarding desired and punishing undesired actions, has been widely used in robotic applications. However, reinforcement learning can be employed in most computerized tasks, such as this study's prediction of visual modalities, in which the learning mode is based on a reward and punishment scheme.

By definition, traditional reinforcement learning, which is modeled as a Markov Decision Process (MDP), can be defined as follows:

$$\boldsymbol{RL} = (\boldsymbol{S}, \boldsymbol{A}, \boldsymbol{P_a}, \boldsymbol{R_a}) \tag{5}$$

where $\boldsymbol{S}$ is the set of possible states the agent can be in. Each state represents a particular configuration of the environment. $\boldsymbol{A}$ defines the set of actions that the agent can take to transition from one state to another. The transition probability $\boldsymbol{P_a}$ , which can be expressed as $\boldsymbol{P_a(s, s') = P_r(s_{t+1} = s' \mid s_t = s,\ a_t = a)}$ defines the probability of transitioning from one state to another when an action is taken. $\boldsymbol{R_a}$ ,which can be expressed as $\boldsymbol{R_a(s, a, s')}$ is the reward function after taking an action $\boldsymbol{a}$ to move from state $\boldsymbol{s}$ to a new state $\boldsymbol{s'}$.

This study focuses on detecting structural damages via the proposed DetectorX framework, which has the adapted EfficientNet-B4 for visual modalities prediction to enhance the percept and robustness of DetectorX to adapt to varying environmental conditions and EfficientDet-D4 for damage detections. The adapted EfficientNet-B4 model in DetectorX is trained based on a type reinforcement learning method, while the adapted EfficientDet-D4 is trained based on supervised learning. However, since the overall objective of DetectorX is the detection of structural damages, a relationship ought to be established between actions (model predictions) of the adapted EfficientNet-B4 and the adapted EfficientDet-D4, or else a reward received by the adapted EfficientNet-B4 may not complement the action (model predictions) of adapted EfficientDet-D4, since the entire DetectorX framework, which serves as an inference model for the micro UAV is being considered as the agent in this reinforcement learning scenario. As a result, event-based reward reinforcement learning (EbRRL), which constrains the actions of both DCNN models in DetectorX as a single action, is proposed.

Event-based reward reinforcement learning is defined as:

$$\boldsymbol{EbRRL} = (\boldsymbol{S^n}, \boldsymbol{A^n}, \boldsymbol{P^n_a}, \boldsymbol{R^n_a}) \tag{6}$$

where $\boldsymbol{S^n}$ is expressed as $\boldsymbol{S + S'}$ in which $\boldsymbol{S}$ represents the classical state space, which is partial and $\boldsymbol{S'}$represent the varied state space, which reveals additional environmental cues via the introduction of visual modalities. Similarly, $\boldsymbol{A^n}$ in Eq. (6) is expressed as $\boldsymbol{A^{xr}_m} + (\boldsymbol{A^{yr}_c} + \boldsymbol{A^{zr}_c})$ where $\boldsymbol{A^{xr}_m}$ represents a parent action with a rewarding weight $\boldsymbol{xr}$ set at 0.4 and $\boldsymbol{A^{yr}_c} + \boldsymbol{A^{zr}_c}$ represents collective child actions with rewarding weights $\boldsymbol{yr}$ and $\boldsymbol{zr}$ set at 0.3, respectively. Detections by the adapted EfficientDet-D4 are the parent actions, and the detection confidence is normalized to the rewarding weight set at 0.4 in $\boldsymbol{A^{xr}_m}$**.** Likewise, the predictions of visual modalities (thermal-like effects and dark theme) by the adapted EfficientNet-B4 denotes the child actions with which the confidence scores are set as the rewarding weights at 0.3, respectively, in $\boldsymbol{A^{yr}_c} + \boldsymbol{A^{zr}_c}$**.** Further, $\boldsymbol{P^n_a}$ in Eq. (6), the collective transition probability is expressed as:

$$\boldsymbol{P^n_a} = \boldsymbol{P_{a^x}(s, s^x)} + \boldsymbol{P'_{a^y+a^z}(s', s'')} \tag{7}$$

where $\boldsymbol{P_{a^x}(s, s^x)} = \boldsymbol{P_r(s_{t+1} = s^x |\ s_t = s,\ a_t = a^x)}$, this is the probability that the next state $\boldsymbol{s_{t+1}}$ will be $\boldsymbol{s^x}$ given the current state $\boldsymbol{s_t}$ is $\boldsymbol{s}$, and the action $\boldsymbol{a_t}$ taken is $\boldsymbol{a^x}$, a parent's action $\boldsymbol{a^x} \in \boldsymbol{A^{xr}_m}$. Also $\boldsymbol{P'_{a^y+a^z}(s', s'')} = \boldsymbol{P_r(s'_{t+1} = s'' |\ s'_t = s',\ a'_t = a^y + a^z)}$**,** this also implies the probability that the next state $\boldsymbol{s'_{t+1}}$ will be $\boldsymbol{s''}$ given the current state $\boldsymbol{s'_t}$ is $\boldsymbol{s'}$, and the action $\boldsymbol{a'_t}$ taken is $\boldsymbol{a^y + a^z}$, thus, the child's actions**,** where $\boldsymbol{a^y} \in \boldsymbol{A^{yr}_c}$**,** and $\boldsymbol{a^z} \in \boldsymbol{A^{zr}_c}$. Finally, $\boldsymbol{R^n_a}$ in Eq. (6), which is expressed as $\boldsymbol{R'_{a^{xr}+a^{yr}+a^{zr}}}$ is the aggregated reward post the parent and child's actions. Based on the definition of the proposed event-based reward reinforcement learning, Eq. (6) can be re-written as:

$$\boldsymbol{EbRRL} = \left(\boldsymbol{S + S'}, \left(\boldsymbol{A^{xr}_m} + \left(\boldsymbol{A^{yr}_c} + \boldsymbol{A^{zr}_c}\right)\right), \left(\boldsymbol{P_{a^x}(s, s^x)} + \boldsymbol{P'_{a^y+a^z}(s', s'')}\right), \boldsymbol{R'_{a^{xr}+a^{yr}+a^{zr}}}\right) \tag{8}$$

In EbRRL, a policy $\boldsymbol{\beta} : \left(\boldsymbol{A^{xr}_m} + \left(\boldsymbol{A^{yr}_c} + \boldsymbol{A^{zr}_c}\right)\right) \longrightarrow [\boldsymbol{0}, \boldsymbol{1}], \boldsymbol{\beta(a_t + a'_t, s + s')} = \boldsymbol{P_r(a_t + a'_t = a^x(a^y + a^z)\ |\ s_t = s + s')}$ is learned to maximize the cumulative reward of the agent, herein DetectorX. Fig. 1(a) gives a graphical insight into the proposed EbRRL.

## 4. Experiment

This section details the dataset, the experimental setup and configuration, and the evaluation metrics used to evaluate the performance of the proposed DetectorX.

### *4.1 Pacific Earthquake Engineering Research (PEER) Hub Image-Net (∅-Net) dataset*

The Pacific Earthquake Engineering Research (PEER) Hub Image-Net dataset, also referred to as ∅-Net dataset [49], comprising varying civil infrastructures, which has eight levels of related tasks, serves as the dataset in this study. Specifically, task 8, thus, damage type, which has four classes, is the focus of this study. The damage type describes the structural damage, thus, irregular semantic vision patterns and complex asymmetrical shapes associated with civil infrastructure. The description of the four classes of damage type is given as follows:

1. Undamaged: This refers to the condition in which there is an absence of any structural deformation.
2. Flexural: This refers to structural damages that manifest in many forms, including but not limited to mold, spalling, and cracks. These damages often occur in a horizontal or vertical pattern on surfaces or at the edges of structures (e.g., pillars, shear walls, and beams).
3. Shear: This refers to structural damage comparable to flexural damage but forms an X, V, or Y asymmetrical shape or a diagonal shape on structures.
4. Combined damage: This refers to structural damage that occurs when there is an uneven pattern in the flexural and shear damage, and the damage is severe.

Fig. 5(a) shows the PEER Hub Image-Net dataset's random samples of task 8, damage type.

### *4.2 Experiment setup and configuration*

The study's experimental setup and evaluation are categorized into three stages: (1) a comparative experiment, (2) a robustness experiment, and (3) a field test experiment. Fig. 5(b) depicts the comparative and robustness experiment setup, and the field test experiment is illustrated in Fig. 5(c). During the field test experiment, a computerized application previously developed by us [46] and a DJI Tello drone, a micro UAV, is used with the inference framework DetectorX for structural damage detection. As illustrated in Fig. 5(c), the computerized application running on an NVIDIA GeForce RTX 3090 with a memory of 32GB receives the video feed of the DJI Tello drone, and DetectorX, which is embedded in the computerized application performs the inference, detections of structural damages. Also, the GamSir ts1 controller is used by a human operator to control the DJI Tello drone. The DJI Tello drone is equipped with a camera, operates on an Intel 14-core processor, has a 5mp capturing rate, has a maximum speed of 8 m/s (meters per second), and has a video streaming quality of 720px. With a wifi repeater (Xiaomi Wifi 2), the DJI Tello drone can fly approximately 200 meters from the human operator.

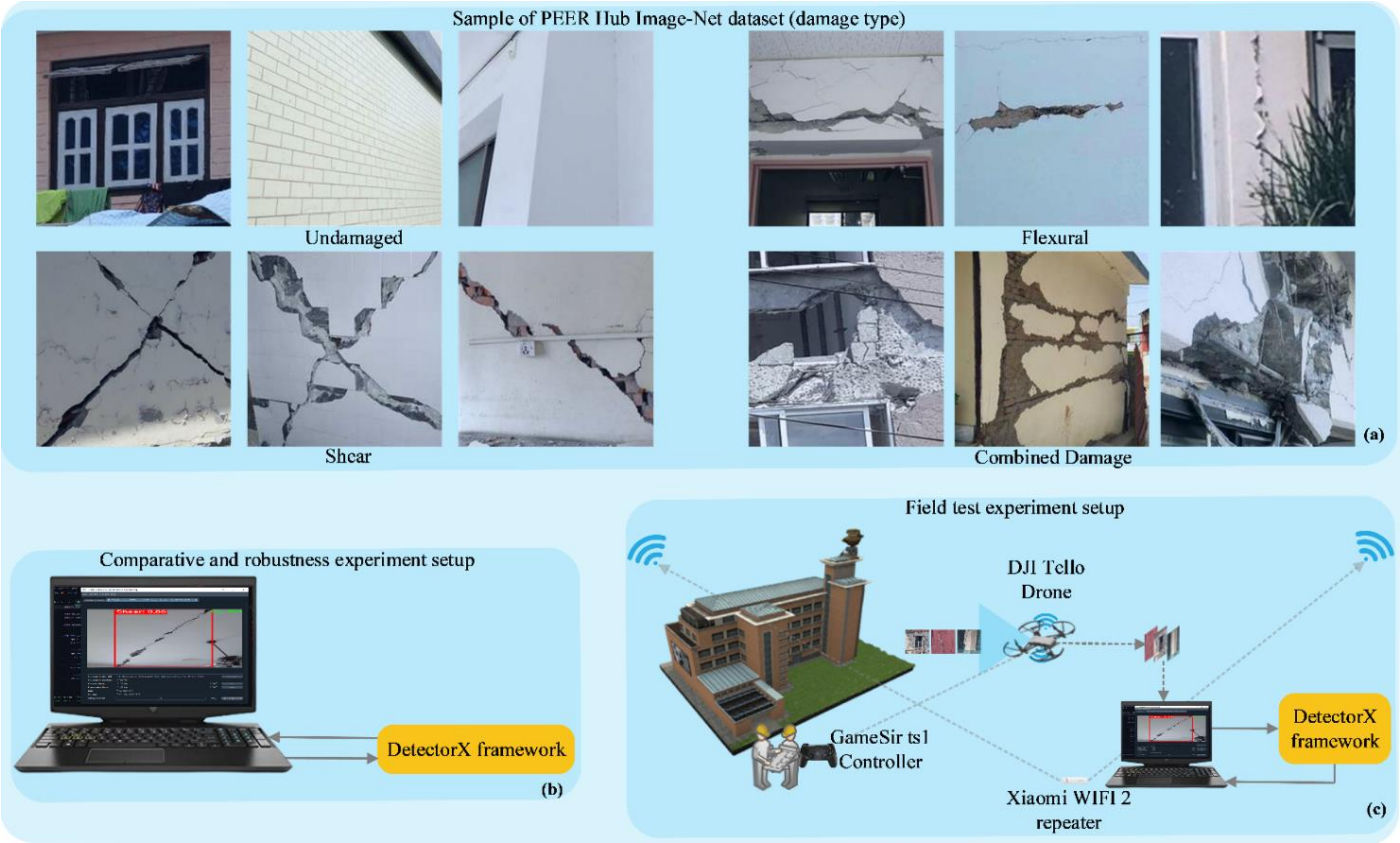

**Fig. 5.** Graphical abstraction of experimental setup and configuration: (a) random samples of training data of the PEER Hub Imagenet (Ø-Net) dataset, specifically damage type, (b) experimental setup for comparative and robustness test, (c) experimental setup for field test.

Regarding the configuration of DetectorX, which has the adapted EfficientNet-B4 and EfficientDet-D4, varying strategies are employed. First, the adapted EfficientNet-B4 is implemented based on EbRRL. Several operations, such as convolutions and spiral pooling, occur, as illustrated in Fig. 4(a), once the data propagates through the various layers of the network. The training hyperparameters of the adapted EfficientNet-B4 include a learning rate = 0.001, batch size = 32, optimizer = Adam, loss function = MSE, discount factor = 0.6, buffer size = 10,000, sampling frequency = 1,000 per training step, and experience replay batch size = 64. On the contrary, the adapted EfficientDet-D4 is implemented based on supervised learning, and the training hyperparameters include a learning rate = 0.001, batch size = 8, epoch = 100, optimizer = Adam, loss function = IOU loss, and a dropout = 0.4. The data resolution set for both models in DetectorX is 380 $\times$ 380 during training and testing. In the ØNet dataset, specifically, the damage type used is 4585 samples, organized into 3000 samples as training data, 1000 samples for validation, and 585 as testing samples.

To train the DetectorX framework, a sequential training mechanism is employed since the adapted EfficientNet-B4 is implemented based on EbRRL, and the adapted EfficientDet-D4 is based on supervised learning. The adapted EfficientNet-B4 training is organized into a train-hold episode; thus, at each episode, there are series of actions (child actions, tuples $T_e$, and $D_t$, refer to Fig. 1 and sub-section 3.4) with collective child rewards, then there is a hold-up for the adapted EfficientDet-D4 to complete an epoch of training to receive a parent reward, which is the prediction score weighted at 0.4. Once the parent's reward is received and aggregated with the collective child reward, as illustrated in Fig. 1, the adapted EfficientNet-B4 reward feedback is completed, and the training sequence switches back to the EfficientNet-B4; this completes one training round of DetectorX.

### *4.3 Evaluation metrics*

The metrics used in this study include precision, recall, F1 score, a normalized confusion matrix, average precision (AP), mean average precision (mAP), mean average recall (mAR), frames per second (fps), and inference time.

The precision, recall, normalized confusion matrix, and F1 Score are computed based on True Positive (TP), False Positive (FP), True Negative (TN), and False Negative (FN). Precision is calculated as $TP\ /\ (TP\ +\ FP)$, where TP represents the number of correctly detected objects, and FP represents the incorrect detections. The recall is computed as $TP\ /\ (TP\ +\ FN)$, where TP represents the number of correctly detected objects, and FN represents the number of missed objects. The F1 score combines precision and recall and is computed as $2\ \times\ (Precision\ \times\ Recall)\ /\ (Precision\ +\ Recall)$. The normalized confusion matrix gives a breakdown of TP, FP, TN, and FN. On the other hand, AP, mAP, and mAR are computed based on the intersection over union (IoU), which measures the overlap between the predicted bounding box and the actual bounding box. The IoU ranges from 0 to 1, with 0 indicating no overlap and 1 indicating a perfect overlap between the two bounding boxes. An IoU threshold of 0.5 determines whether a predicted bounding box is sufficiently close to the ground truth bounding box to be considered a valid detection.

## 5. Results and discussion

In all three evaluation categories, (1) the comparative experiment, (2) the robustness experiment, and (3) the field test experiment, DetectorX is compared with some current (≥ 2019) object detectors (EfficientDet-D5 [48], YOLOX-m [50], and YOLOv5-m [51]) within similar time frame of release of the base model, which DetectorX is built upon and relatively old (≤ 2018), yet in-service state-of-the-art detectors, including SSD500 [52], and Faster-R-CNN [53]. Also, to ensure a fair comparison, varying augmentation, including saturation, flips, hue, and random cropping, were added to the data on which the compared detectors were trained. Additionally, two ablated versions of DetectorX, namely DetectorX-SB and DetectorX-SP, are included in the comparators. DetectorX-SB is the DetectorX framework, excluding the stem block, which induces dynamic visual modalities, and DetectorX-SP is DetectorX without the spiral pooling feature. Aside from the ablated versions of DetectorX, the selected comparators' training follows a similar training process outlined in the respective studies.

### *5.1 Comparative evaluation and analysis*

This section is organized into quantitative analysis and normalized confusion matrix analysis. The quantitative analysis comprises precision, recall, F1 score, AP, mAP, mAR, FPS, inference time, and Floating Point Operations (FLOPs).

#### *5.1.1 Quantitative analysis*

This section analyses and discusses the tabular results in Table 2, supplemented by Fig. 6 for graphical insight. The results give valuable insights into the performance of DetectorX in structural damage detection.

From the quantitative results in Table 2, DetectorX demonstrates better performance across multiple metrics, including precision (0.88), recall (0.84), and F1 score (0.86), highlighting its ability to detect structural damages with a high degree of accuracy compared to the other detectors in the study. The high precision and recall metrics score indicates DetectorX's ability to minimize false positives and detect actual objects of interest effectively. Also, DetectorX's high F1 score signifies a good trade-off between the precision and recall metrics. Furthermore, DetectorX achieves a satisfactory AP of 0.91, indicating its proficiency in accurately classifying and localizing damaged regions within images compared to the competing detectors. The higher AP results attained by DetectorX indicate its superior capabilities in accurate localization and correct classification of class instances. Regarding the mAP and mAR, taken across varying detection thresholds, which also provide comprehensive insight regarding the detector's performance, it is evident that DetectorX stands out compared to the detectors in this study, as seen in Table 2. DetectorX mAP and mAR fall within the seventieth percentile, indicating that DetectorX maintains a high accuracy and consistency across varying detection thresholds (0.5, 0.55, …, 0.95). Lastly, although DetectorX does not outperform all the detectors regarding the inference time and fps, an inference time hovering around 23ms in the three repetitions of the inference time test, as seen in Fig. 6 (d), and an fps of approximately 35ms in Table 2 is conducive for some real-time or near-real-time applications, respectively. Additionally, DetectorX exhibits quite high FLOPs of 89.2B, indicating a substantial computational load. From the base models, the two additional modules, the stem block and the spiral pooling mechanism contribute around 30B FLOPs. Despite a quite high FLOP, DetectorX's overall performance across various metrics, including precision, recall, F1 score, AP, mAP, and mAR, is satisfactory.

**Table 2**
Quantitative comparison of DetectorX, variants of DetectorX, and SOTA detectors on the PEER Hub Image-Net damage type dataset.

| Model | Precision | Recall | F1 Score | AP | mAP | mAR | FPS | Inference time | FLOPs |
|---|---|---|---|---|---|---|---|---|---|
| DetectorX | 0.88 | 0.84 | 0.86 | 0.91 | 0.76 | 0.73 | 35.16 | 23.32 | 89.2B |
| DetectorX-SB | 0.81 | 0.79 | 0.80 | 0.80 | 0.69 | 0.67 | 39.23 | 20.57 | 66.5B |
| DetectorX-SP | 0.85 | 0.81 | 0.83 | 0.86 | 0.72 | 0.70 | 41.41 | 19.09 | 63.4B |
| EfficientDet-D5 | 0.83 | 0.80 | 0.81 | 0.84 | 0.70 | 0.69 | 38.04 | 17.34 | 135B |
| YOLOX-m | 0.86 | 0.82 | 0.84 | 0.89 | 0.73 | 0.71 | 80.21 | 11.33 | 73.8B |
| YOLOv5-m | 0.79 | 0.77 | 0.78 | 0.80 | 0.68 | 0.66 | 119.03 | 4.30 | 39.3B |
| SSD500 | 0.70 | 0.67 | 0.69 | 0.74 | 0.61 | 0.59 | 19.03 | 55.12 | 212B |
| Faster-R-CNN | 0.71 | 0.69 | 0.70 | 0.76 | 0.64 | 0.60 | 10.14 | 98.56 | 886B |

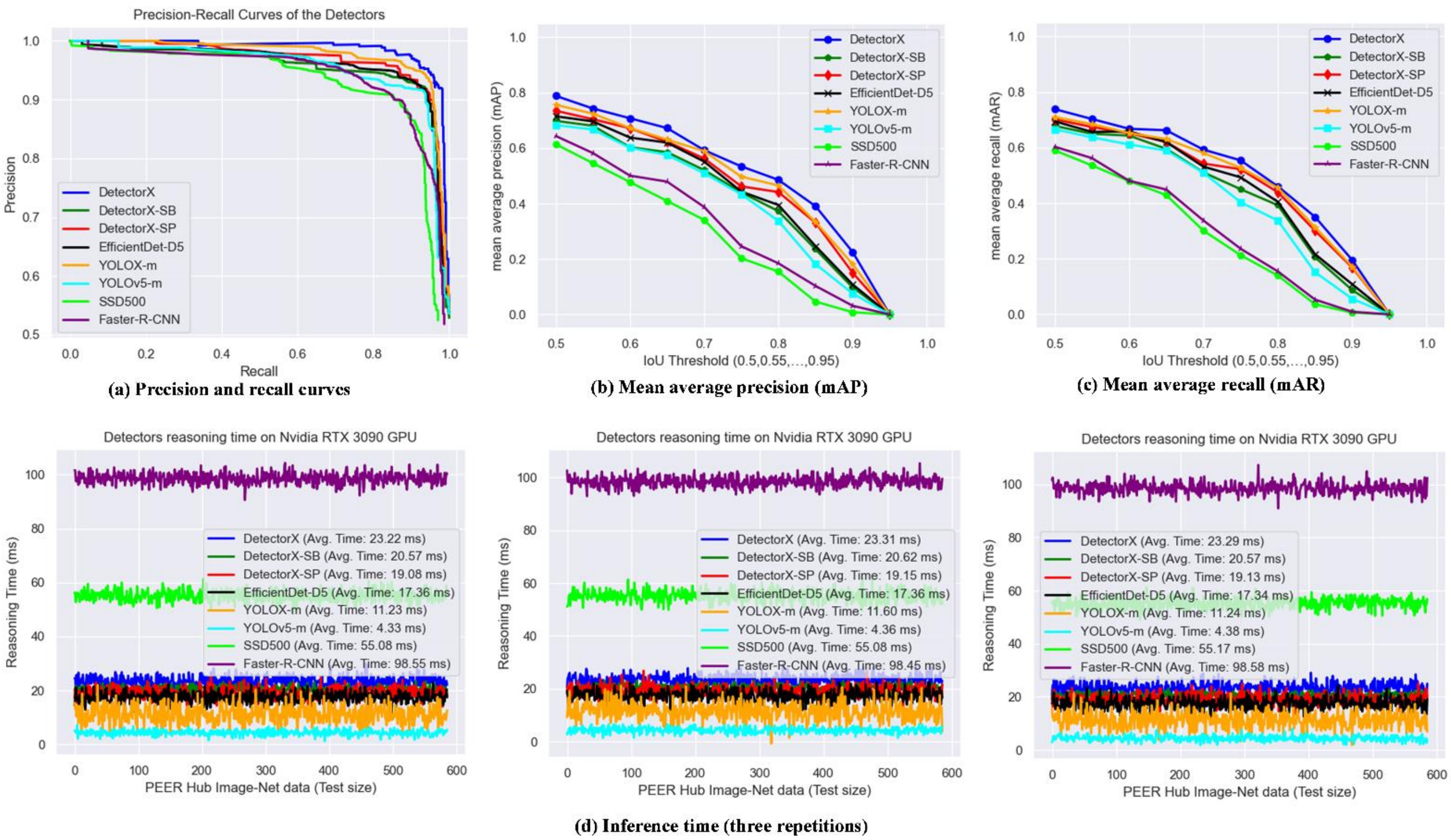

**Fig. 6.** The graphical comparison of DetectorX, SOTA detectors, and the ablated DetectorX versions regarding (a) precision and recall, (b) mean average precision, (c) mean average recall, and (d) the three repetitions of the inference time test.

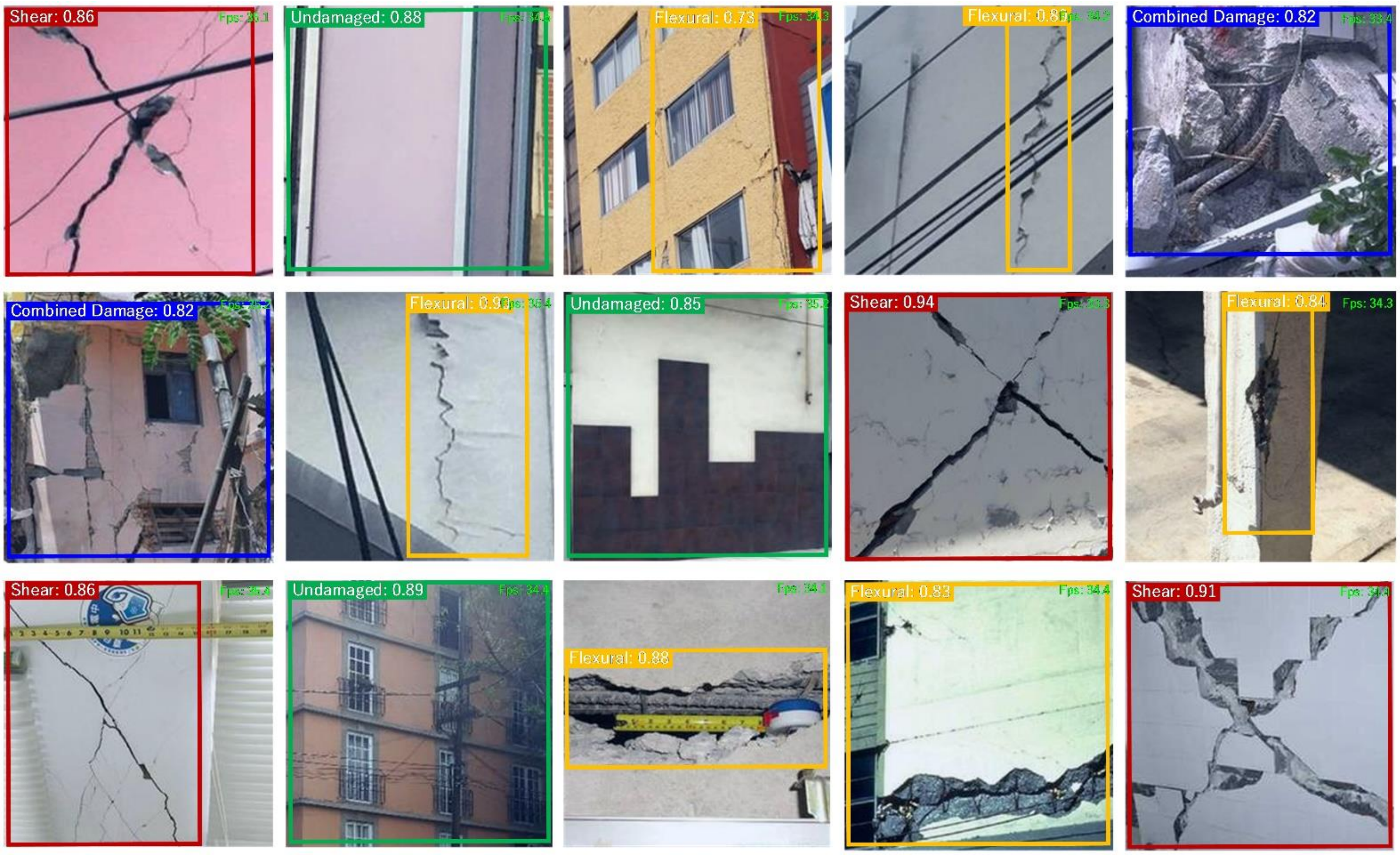

**Fig. 7.** Varying damage type detection by DetectorX on PEER Hub ImageNet dataset.

The success of DetectorX is attributed to the innovative design of the framework, which incorporates two adapted DCNN models, modified EfficientNet-B4 and EfficientDet-D4. Unlike traditional DCNN models, the modified EfficientNet-B4 in DetectorX utilizes the proposed event-based reward reinforcement learning to induce dynamic visual modalities, creating two distinct visual modalities alongside the RGB input data. Consequently, DetectorX operates on three visual modalities, enhancing its percepts, robustness, and adaptability to varying scenes. The framework's effectiveness is further enhanced by introducing spiral pooling, an online image augmentation technique. Spiral pooling capitalizes on the inherent structure of spiral patterns, enabling the augmentation of feature maps during training. This technique increases the variation of the feature representations and spatial relationships, contributing to the satisfactory performance of DetectorX.

Comparing DetectorX with its variants, DetectorX-SB and DetectorX-SP in Table 2 and the supplemented Fig. 6, it is evident that DetectorX retains its superiority regarding precision, recall, F1 score, AP, mAP, and mAR. The variants attained slightly better performance than DetectorX on the inference time and the fps, as seen in Table 2; this was expected since the variants' model parameters are less than DetectorX. Also, DetectorX-SB and DetectorX-SP demonstrate lower FLOPs than DetectorX (66.5B and 63.4B, respectively). The slight decrease in FLOPs for the variants is because of the fewer model parameters. While the variants show a marginal improvement in inference time and FPS, DetectorX retains its superiority in key metrics like precision, recall, and F1 score. The comparison between the variants and DetectorX indicates that the induction of the dynamic visual modality along with the RGB input impacts the performance of DetectorX more than that of the spiral pooling. The dynamic visual modality offers two additional visual insights into the image environments, while the spiral pooling is much centered on the orientation of the features within the image. From Table 2, across all the metrics except the inference time and the fps, DetectorX without the stem block, the variant DetectorX-SB experiences a drop of 5% in performance in most of the metrics. In contrast, DetectorX, without the spiral pooling, thus, Detector-SP, experiences a drop of 3% in most of the metrics. These results emphasize the importance of the proposed innovative techniques and the contributions of the two proposed modules, event-based reward reinforcement learning and spiral pooling, in advancing the field of structural damage detection and beyond.

In contrast to DetectorX and its variants, the other state-of-the-art (SOTA) object detection models, including EfficientDet-D5, YOLOX-m, YOLOv5-m, SSD500, and Faster-R-CNN, show varying levels of promising performances. Among the SOTA detectors, YOLOX-m outperforms all the SOTA detectors in addition to DetectorX variants; however, the performance between YOLOX-m and DetectorX-SP are competitive, usually $\pm 1$, regarding precision, recall, F1 score, AP, mAP, and mAR metrics. Following the performance of YOLOX-m is the EfficientDet-D5, which attains promising performances in most metrics but lags behind DetectorX and slightly behind the variant DetectorX-SP. Next is the YOLOv5-m, which records precision and recall within the seventieth percentile, a mAP, and mAR within the sixtieth percentile, as seen in Table 2. Although the YOLOv5-m attains some good performances, it trails the performances of DetectorX and both variants, DetectorX-SP and DetectorX-SB. Among the SOTA detectors, the Faster-R-CNN and SDD500 adduce the lowest performance in detecting damage types regarding all the evaluation metrics utilized. Although the Faster-R-CNN and SDD500 have the lowest performance, the Faster-R-CNN is slightly better than the SDD500. Also, among the SOTA detectors, the YOLO-based models attained the best results regarding inference time and fps in the study, as seen in Table 2; this is usually the strength of YOLO-based models. Also, EfficientDet-D5, YOLOX-m, YOLOv5-m, SSD500, and Faster-R-CNN exhibit a wide range of FLOPs. EfficientDet-D5 attained high FLOPs of 135B, reflecting its computational complexity. YOLOX-m and YOLOv5-m have FLOPs of 73.8B and 39.3B, respectively, showing their efficiency in terms of computational load. In contrast, Faster-R-CNN and SSD500 record the highest FLOPs, which are 886B and 212B, respectively. The FLOPs, when considered alongside other metrics, contribute to the overall understanding of the models' trade-offs between accuracy and computational demands. While FLOPs provide valuable insights into the computational load of the models, they should be considered in conjunction with the other metrics to comprehensively assess a model's performance. In addition to Table 2 and Fig. 6, Fig. 7 shows some random samples of detections of the damage type by DetectorX.

In summary, the results demonstrate that DetectorX offers a well-rounded performance showcasing high precision, recall, F1 score, AP, mAP, and mAR, as well as a competitive inference and fps time. These results underscore its potential for a wide range of applications requiring precise and reliable object detection. However, it is crucial to consider the application's specific requirements, such as the trade-offs between accuracy and speed. Ultimately, these findings potentially contribute significantly to the ongoing advancements in object detection techniques and provide valuable guidance to researchers and practitioners.

#### *5.1.2 Normalized confusion matrix analysis*

The normalized confusion matrix provides additional insight into the classification performance of the object detectors; thus, it helps to evaluate the detectors' ability to classify true positives accurately (correctly detected structural damages), true negatives (correctly detected non-damages), false positives (incorrectly detected damages), and false negatives (missed structural damages).

DetectorX exhibits the highest precision, recall, and F1 score among the detectors, indicating that it correctly detects structural damages and minimizes false alarms. The normalized confusion matrix in Fig. 8 reinforces the following findings:

1. High True Positive (TP) rate: In the confusion matrix, the upper-left quadrant represents true positives, the most critical aspect of structural damage detection. DetectorX has a high TP rate, indicating its effectiveness in correctly classifying and localizing structural damages compared to the other detectors; this is crucial in real-world applications where missing damage can have severe consequences. Among the comparators, DetectorX variants, EfficientDet-D5 and YOLOX-m attain a good TP rate but trail that of DetectorX. Also, YOLOv5-m achieved a higher TP but less than DetectorX, the variants, and the other SOTA models except for SSD500 and Faster-R-CNN, which perform the least.

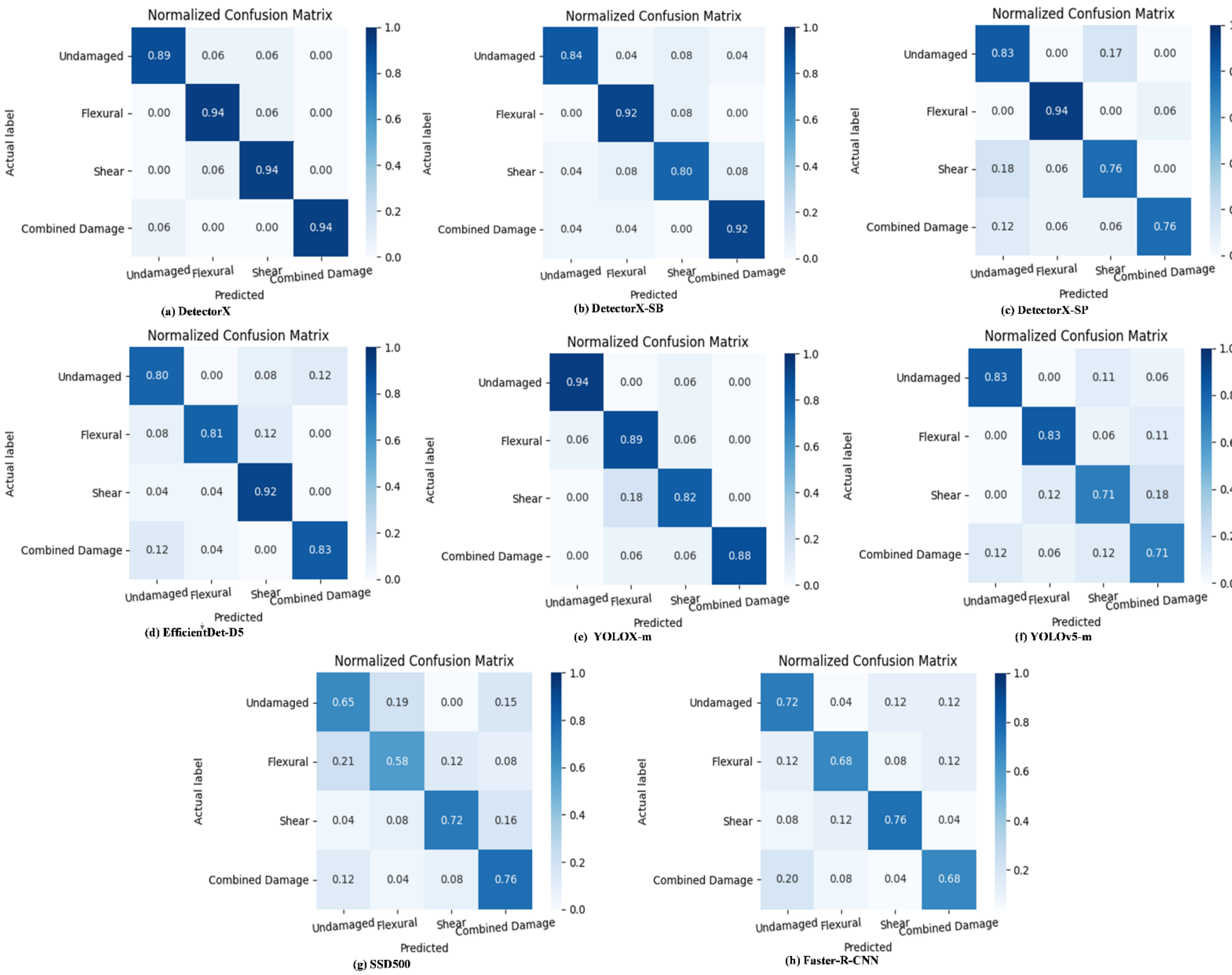

Fig. 8. A normalized confusion matrix comparison between DetectorX, variants of DetectorX, and SOTA detectors.

2. Low False Positive (FP) rate: DetectorX's precision score, being the highest among the detectors, indicates its lowest FP rate; this means that when DetectorX identifies structural damage, it is highly likely to be a true positive rather than a false alarm compared to that of the other detectors. It is worth mentioning that DetectorX variants, YOLOX-m, EfficientDet-D5, and, to an extent, YOLOv5-m attained a varying low FP but not lower than DetectorX. Low FP rates are essential in preventing unnecessary maintenance post-structural inspection.
3. High True Negative (TN) rate: The lower-right quadrant of each detector's normalized confusion matrix represents true negatives. DetectorX's ability to accurately classify non-damages as negatives contributes to a high TN rate, as seen in Fig. 8 (a), compared to the others, as seen in Fig. 8 (b) to (h); this is crucial in structural health monitoring since some of the inspected parts of the civil infrastructure potentially may not be damaged.
4. Low False Negative (FN) rate: The FN rate, representing missed detections, is exceptionally low for DetectorX compared to some comparators; this indicates that DetectorX rarely misses structural damages, which is vital for ensuring the safety and integrity of structures. Minimizing FN rates is a top priority in structural damage detection. Aside from DetectorX, the YOLO-based detectors, EfficientDet-D5, and the variants of DetectorX also attained varying low FN but not lower than DetectorX. On the contrary, such can not be said about the SDD500 and Faster-R-CNN detectors.
5. Overall balanced performance: The balanced performance across TP, FP, TN, and FN rates in DetectorX's confusion matrix reflects its robustness in detecting structural damages while maintaining a low false alarm rate compared to all the detectors in this study. This balance is particularly important in safety-critical applications, where precision and recall must be high, and DetectorX exhibits such characteristics. Although YOLOX-m, DeetctorX-SP, EfficientDet-D5, and DetectorX-SB adduced a precision and a recall above the 80% mark excerpt for the recall of DetectorX-SB, which is 79%, DetectorX attained a much better performance.

In conclusion, the normalized confusion matrix reaffirms the superior performance of DetectorX in structural damage detection, which is attributed to the framework's design. Its high precision, recall, and F1 score, combined with the characteristics observed in the confusion matrix, demonstrate its effectiveness in correctly detecting damages while minimizing false alarms. DetectorX's ability to achieve a balanced performance in TP, FP, TN, and FN rates makes it an additional model of choice for critical applications where safety and accuracy are paramount.

### *5.2 Robustness evaluation and analysis*

#### *5.2.1 Low to high intensities evaluation and analysis*

Additional tests are conducted to assess the robustness of the detectors, which is critical to evaluating the reliability and performance of the detection models, especially in real-world scenarios where environmental conditions can vary significantly. The robustness of structural damage detection of DetectorX, its variants, and the SOTA detectors are assessed under four challenging conditions: (1) blurring, (2) illumination, (3) noise, and (4) fog each at low, mild, and high intensities. The intensity scales are tabulated in Table 3. Precision and recall metrics are utilized to assess the performances of DetectorX and the compared detectors, and the results are visualized in Fig. 9, together with some detection results of DetectorX.

As seen in Fig. 9 (a), DetectorX demonstrates better resilience across varying blurring intensities than the compared detectors. It maintains high precision and recall at low blurring intensity, excelling at detecting structural damages in slightly blurred images. As blurring intensifies to mild levels, DetectorX outperforms other detection models with slightly lower but consistent high precision and competitive recall, making it suitable for scenarios with moderate image distortions. Even at high blurring intensities, DetectorX maintains commendable precision and recall, making it reliable in challenging conditions. In contrast, except for YOLOX-m, which shares similar resilient but lower performance to DetectorX, the other SOTA detection models and the variants of DetectorX exhibit varying sensitivity to blurring, with SSD500 and Faster-R-CNN experiencing significant drops in precision and recall when the intensities increase. DetectorX's high performance across blurring levels underscores its robustness and suitability for real-time or near real-time applications where image quality may be compromised due to vibrations or mild weather-induced distortions.

Further, in illumination conditions, in a more challenging environment, DetectorX maintains fairly high precision from 70% to 81% and recall of 68% to 78%, as seen in Fig. 9 (b), enabling it to detect structural damages with minimal false. Despite slightly lower recall and precision than in well-lit conditions and even in blurring and noisy conditions, DetectorX detects significant damages, enhancing safety and structural integrity. Under mild illumination, DetectorX maintains a slightly lower but high precision and moderate recall, detecting damages and ensuring reliability in diverse lighting environments except for high illumination conditions where precision and recall are 70% and 68%, respectively, which arguably compromises the reliability attribute of DetectorX. Among the comparators, the best-performing detector, the YOLOX-m, adduced a precision of 68% and a recall of 66% in higher illumination conditions, solidifying that DetectorX is much better even in intense conditions. In contrast, the least performed detector, SSD500, attained a precision of 53% and recall of 50% in highly intense illumination conditions.

Also, in assessing the impact of noise intensity on precision and recall, DetectorX consistently demonstrates robust performance.

**Table 3**
The intensity levels of the challenging environmental conditions utilized in the robustness test.

| Low intensity | Mild intensity | High intensity |
|---|---|---|
| | Blurring condition | |
| Blurring kernel_size = 10, 10 | Blurring kernel_size = 15, 15 | Blurring kernel_size = 20, 20 |
| | Illumination condition | |
| Incandescence_factor = 0.9<br>Luminescence_factor = 0.6 | Incandescence_factor = 1.1<br>Luminescence_factor = 0.6 | Incandescence_factor = 1.3<br>Luminescence_factor = 0.6 |
| | Noise condition | |
| Mean = 2, standard deviation for Gaussian noise = 4 | Mean = 2, standard deviation for Gaussian noise = 8 | Mean = 2, standard deviation for Gaussian noise =12 |
| | Fog condition | |
| Fog blending intensity = 0.3 | Fog blending intensity = 0.6 | Fog blending intensity = 0.9 |

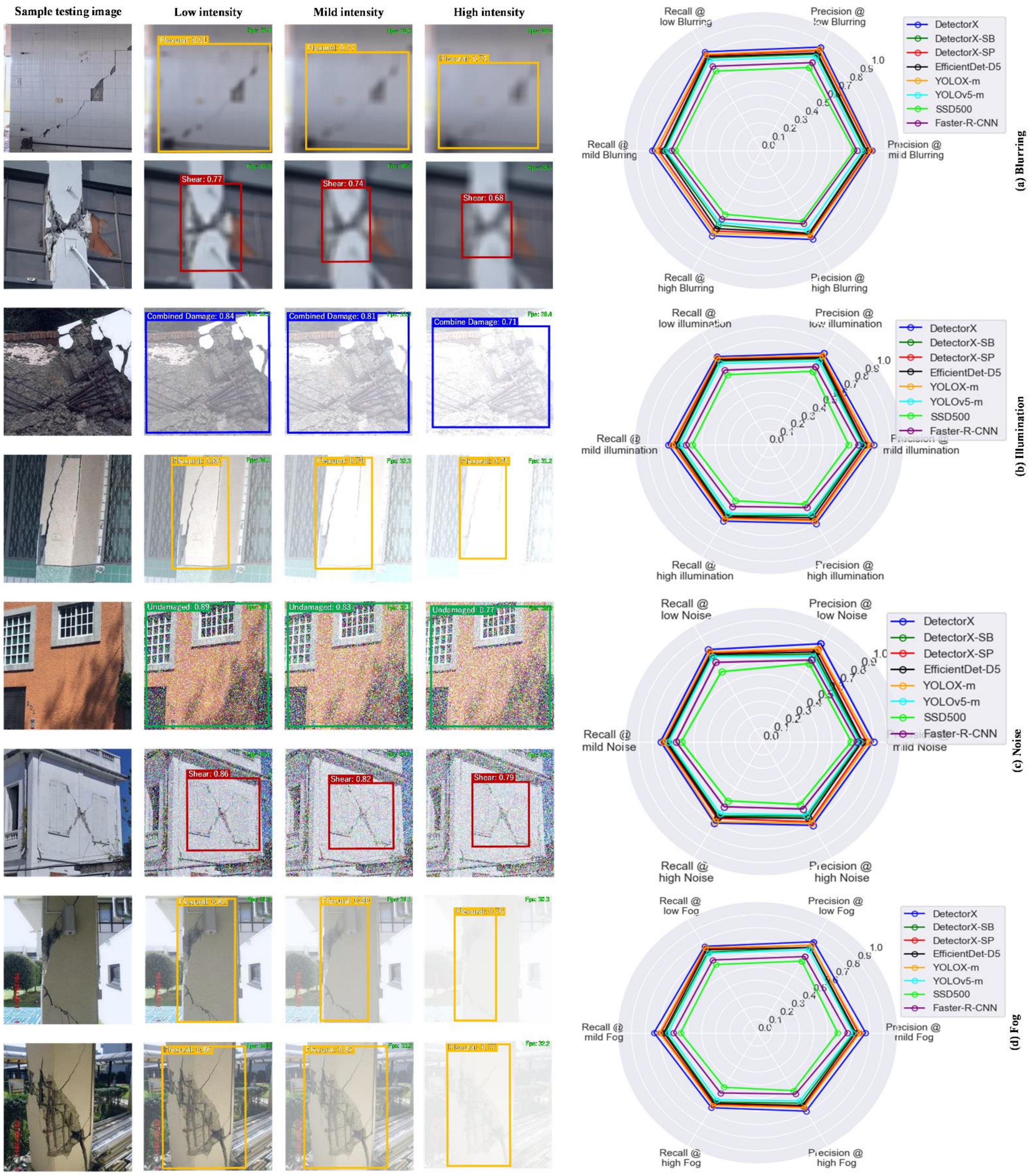

**Fig. 9.** Robustness test under varying image environment conditions for DetectorX, variants of DetectorX, and SOTA detectors: (a) blurring conditions from low to high intensities, (b) illumination conditions from low to high intensities, (c) noise conditions from low to high intensities, and (d) fog conditions from low to high intensities.

Under low noise intensity, it maintains a precision of 84% and a recall of 79% compared to the best comparator, YOLOX-m, which recorded a precision of 80% and a recall of 77%. In the mild noise scenario, DetectorX remains resilient; however, in the intense noise condition, DetectorX adduced a precision of 72% and a recall of 70%, which shows a moderate drop in performance. In contrast, YOLOX-m, the best-compared detector, adduced a precision of 70% and a recall of 69% in highly intense conditions. On the other hand, the least performed detector, SSD500, attained precision and recall scores of 54% and 51%, respectively, in highly intense conditions. Overall, DetectorX's reliability across noise levels positions it as a resilient detector in low to mild noise environments, while it is competitive in extremely noisy conditions. These findings emphasize its practicality and potential for enhancing safety and efficiency in critical infrastructure assessments.

Lastly, in the fourth robustness test, similar results attained during the illumination test are observed under fog conditions. The fog and illumination are the most challenging environments among the four varying robustness environments. In the fog conditions, DetectorX attained commendable precision and recall, mostly in low and moderate intensity. In high intensity, DetectorX performance was reduced regarding precision and recall. In addition, a secondary observation is a reduction in the detection area (bounding box) when the intensity is high. Among the compared detectors, YOLOX-m attained close performance to DetectorX, while the variants of DetectorX adduced varying promising performances in low to mild intensities. It must be noted that state-of-the-art detectors, including EfficientDet-D5 and YOLOv5-m, also attained satisfactory results in the low to mild intensities in the fog conditions but trailed the performance attained in other conditions, such as noise and illumination. In contrast, the two least-performing detectors, the SSD500 and the Faster-R-CNN detectors, continued to underperform among the detectors.

In conclusion, the robustness of structural damage detection demonstrated by DetectorX makes it promising for vital practical applications in diverse real-world environments. While DetectorX is designed to handle variations in image conditions, extreme blurring, illumination changes, noise or fog levels can reduce the performance, as discussed in sub-section 5.2.2.

### *5.2.2 Extreme intensities evaluation and analysis*

The robustness evaluation extends to extreme conditions, exploring the impact of extreme intensities in blurring, noise, and fog on the structural damage detection performance of DetectorX and the compared detectors. The blurring kernel is set at 40 by 40, while the noise Mean remains at 2, and the standard deviation for Gaussian noise is set at 50. Lastly, the fog intensity is set as 0.975. This examination is crucial for understanding the model's limitations and potential areas for improvement in challenging scenarios.

In extreme conditions, a significant drop in performance is experienced by all the detectors in the study. Although there is a drop in performance, DetectorX exhibits a commendable robustness considering how intense the extreme conditions are. DetectorX maintained precision and recall between 44% and 38%, showcasing its resilience to extreme blurring, noise and fog conditions. In addition, a secondary observation was that structural damages on infrastructure at the object or structural level were challenging to detect in extreme conditions compared to the pixel-level objects of interest. Among the three extreme conditions (bur, noise, and fog), the extreme fog proved to be the most challenging. DetectorX outperforms the variants as well as the comparators, reinforcing its practicality in environments with varying extreme levels of intensity.

On the other hand, the two ablated versions of DetectorX also experienced significant drops in performance, attaining precision and recalls between 41 and 35. Similarly, the other state-of-the-art detectors also experienced varying significant drops in performance, as seen in Fig. 10. Among the comparable detectors, YOLOX-m attained the best results with precision and recalls between 42% and 37%. Although the compared detectors experienced significant drops, detectors including EfficientDet-D5, YOLOX-m, and YOLOv5-m can still be used in extreme conditions since they can make some detections. In contrast, the SSD500 and the Faster-R-CNN detectors exhibited extremely significant drops in performance, recording precision and recall between 26% and 25%. These low precision and recall results for the SSD500 and the Faster-R-CNN detectors make them unsuitable in extreme conditions.

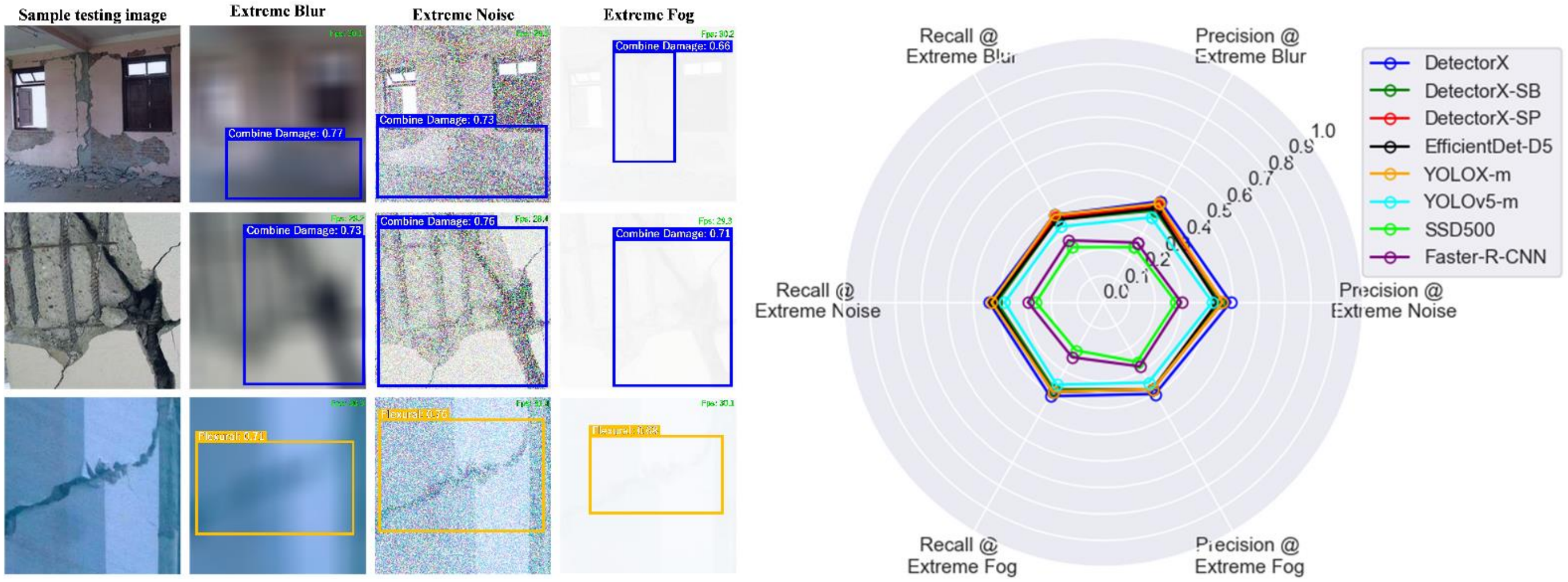

**Fig. 10.** Extreme robustness test under blurring, noise, and fog conditions for DetectorX, variants of DetectorX, and SOTA detectors.

In summary, DetectorX demonstrates a good level of robustness across extreme blurring, noise, and fog conditions. However, extreme fog conditions present severe challenges, indicating the need for further advancements in handling such scenarios. The results underscore DetectorX's potential for real-world applications with varying environmental conditions while also providing insights into areas that could benefit from future research and model enhancements.

### *5.3 Field-test evaluation and analysis*

The field-test experiment assesses DetectorX and the compared detectors' performance in real-world settings, and it is grouped into (1) micro drone speed to detection performance and (2) distance to detection performance. These assessments provide practical insights into DetectorX's suitability for applications requiring accurate damage assessment.

#### *5.3.1 Micro drone speed-to-detection evaluation and analysis*

The impact of the speed of the micro drone on detection performance is evaluated to gain further insights into the suitability of DetectorX and the other detection models. The speed-to-detection performance examines the precision and recall results obtained for DetectorX and the other detection models at varying speeds: approximately 4-5 meters per second and 7-8 meters per second, as the micro drone flies horizontally at a distance of approximately 30 meters across a wall stretched slightly over 60 meters. Since the speed of the micro drone is manually controlled, there are minor drops in speed, as seen in Fig. 11; however, the best effort was ensured to fly at 4-5 and 7-8 meters per second, respectively.

Some detection models demonstrated promising performances at the speed testing of 4-5 meters per second. DetectorX emerged as the most promising detector, with commendable precision and recall scores of 0.7886 and 0.7386, respectively. These results emphasize DetectorX's ability to maintain a good level of accuracy in scenarios involving a moderate-dynamic environment. DetectorX's consistent performance underscores its robustness in dynamic environments where accurate object detection at high speeds is imperative. DetectorX-SP, while exhibiting slightly lower precision and recall compared to DetectorX, still attained acceptable performance with a precision of 0.7349 and a recall of 0.7019. Similarly, the best-compared detector to DetectorX, YOLOX-m, attained a precision of 0.7563 and a recall of 0.7102. Likewise, the EfficientDet-D5 attained similar performances but were slightly lower than YOLOX-m. Also, the variant DetectorX-SB and the SOTA model YOLOv5-m recorded an average precision hovering around 0.69, with a recall of 0.67. In contrast, SSD500 and Faster-R-CNN exhibit lower precision and recall scores than other detectors, as seen in Fig. 11 (a); these results suggest that SSD500 and Faster-R-CNN are arguably less suitable for moderate-speed applications.

Further, when the micro drone speed was 7-8 meters per second, the performance of detection models was put to a more demanding test as all of them experienced reductions in precision and recall, as evident in Fig. 11 (b). DetectorX recorded commendable precision and recall scores of 0.7464 and 0.7046, respectively; this reaffirms DetectorX's capacity to detect damages in a fast-moving environment with some accuracy, particularly in safety-critical scenarios. In this challenging test, YOLOX-m precision was 0.7213, with a recall falling to the sixty percentile; thus, 0.6873. Similarly, DetectorX-SP and EfficientDet-D5 attained precision and recall, which are acceptable due to the rigorous nature of the test. Also, DetectorX-SB and YOLOv5-m precisions hover around 0.65, while their recalls are around 0.62. The least performed detectors, SSD500 and Faster-R-CNN, exhibited lower precision and recall scores at the higher speed testing rates. Their performance dwindled when faced with fast-moving environments, suggesting that they may be less suitable for high-speed applications where maintaining precision and recall at a high level is essential.

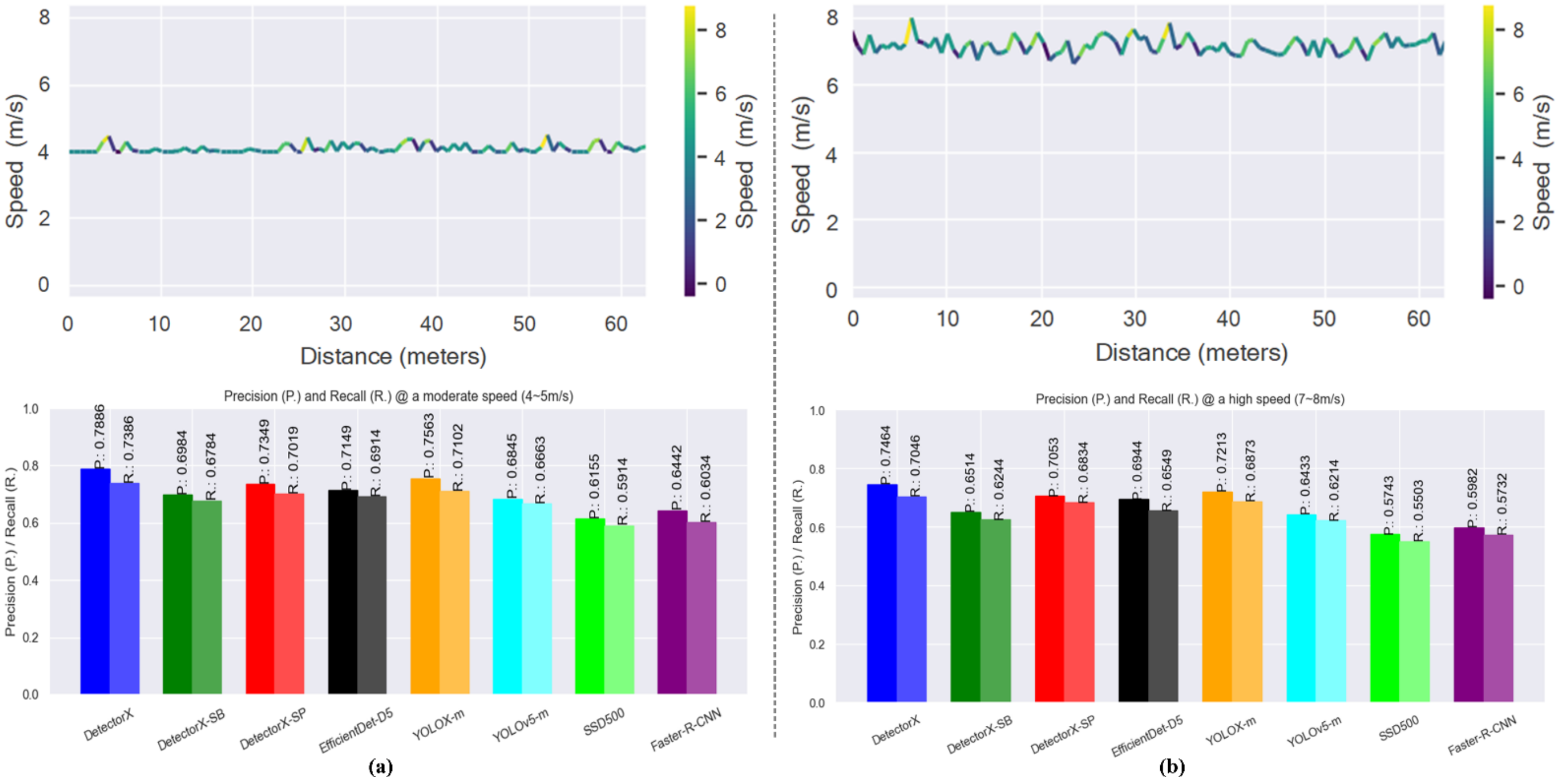

**Fig. 11.** Mirco drone speed-to-detection performance of DetectorX and competing detection models at varying speeds: (a) approximate speed 4-5m/s, and (b) approximate speed 7-8m/s.

Overall, the micro drone speed-to-detection evaluation highlights the diverse performance of detectors in scenarios with varying speed testing rates. DetectorX emerges as the best detector among the contenders, including YOLOX-m, DetectorX-SP, and EfficientDet-D5. While DetectorX-SB and YOLOv5-m showed some promising performances, they are categorized as moderately performed detectors, whereas Faster-R-CNN and SSD500 performed the least. Lastly, the reader should note similar but varying performances may be attained when a different type of micro-drone with similar or better qualities (5 mega-pixel, 720px) is used. In addition, varying flight dynamics exhibited by variant micro-drones may also result in similar but varying performances attained in this study when a varying micro-drone is used.

### *5.3.2 Distance to detection evaluation and analysis*

An additional test was conducted to evaluate the performance of DetectorX and the compared detectors at different distances from the target area, specifically at distances of 15, 30, and 50 meters, respectively. During this test, the micro drone flies and hovers once a desired altitude is reached. This evaluation focused on assessing the ability of the detectors to accurately detect objects at varying distances, which is critical in applications like structural health monitoring. The precision and recall results for each detector at the testing distances are visualized in Fig. 12.

At 15 meters from the target area, all the detectors exhibited satisfactory performance except for SSD500 and Faster-R-CNN, which are average performances. DetectorX outperformed all the detectors with a precision of 0.8433 and a recall of 0.8303. YOLOX-m and DetectorX-SP followed closely with precisions of 0.8155 and 0.7953 and recalls of 0.8034 and 0.7751, respectively, indicating their competitive nature. Also, EfficientDet-D5, DetectorX-SB, and YOLOv5-m showed satisfactory performances, as seen in Fig 11 (a), with precisions above 0.75 and recall scores above 0.71. Even the detectors with slightly lower scores, such as SSD500 and Faster-R-CNN, displayed a normal and acceptable performance at this distance, making them arguably viable options for scenarios where precision and recall are critical within a 15-meter distance to inspection targets.

As the distance increased to 30 meters from the target area, DetectorX and other detectors attained reliable results. DetectorX maintained satisfactory precision and recall with values of 0.8204 and 0.8113, respectively, highlighting its consistency in object detection at this intermediate distance. YOLOX-m and DetectorX-SP, as expected, also performed well, with precision scores of 0.7613 and 0.7403 and recall scores of 0.7904 and 0.7751, respectively, reaffirming their competitive nature in this study. Like in the 15-meter to target area test, EfficientDet-D5, DetectorX-SB, and YOLOv5-m showcased above-average precision and recall scores, further emphasizing their suitability for scenarios where objects must be detected within a 30-meter range. However, the Faster-R-CNN and SSD500 detectors experience some challenges as their precision and recall decline, as seen in Fig. 12 (b).

Lastly, DetectorX and some competitive detectors, including YOLOX-m and a variant, DetectorX-SP, displayed resilient performance at 50 meters from the target area, as seen in Fig. 12 (c), with some variation in precision and recall scores. DetectorX maintained commendable precision and recall values of 0.7813 and 0.7641, respectively, underlining its ability to detect objects with some accuracy even at a range of 50 meters. DetectorX-SP and YOLOX-m attained good results with precision and recall scores ranging between 0.73 and 0.75 and 0.71 and 0.72, respectively. Further, EfficientDet-D5 and DetectorX-SB precision and recall scores stayed within the 70% mark while YOLOv5-m dropped to the 60% range. At the 50-meter distance, SSD500 and Faster-R-CNN continued to perform less with lower precision and recall scores compared to the 30-meter distance test.

In conclusion, the performance of DetectorX and the other detection models was evaluated across different distances, namely 15, 30, and 50 meters from the target area. Even though DetectorX and some competitive detectors performed satisfactorily across the three varying distances, their precision and recall reduced as the distances increased; these findings indicate that DetectorX is a better detector among the other competing detectors for a wide range of detection scenarios, offering promising and reliable performance across varying distances.

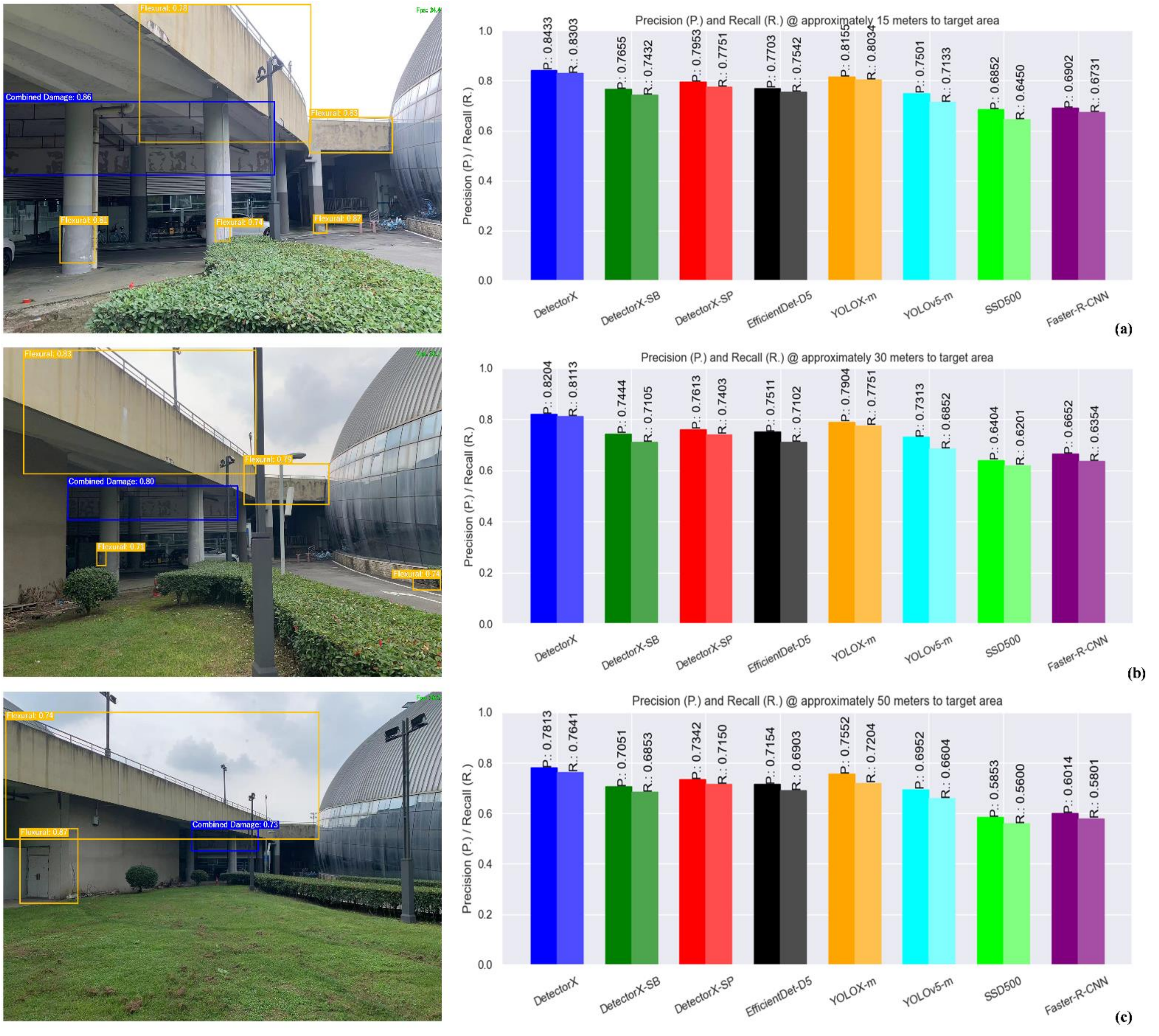

**Fig. 12.** Detection performance at different distances regarding precision and recall results for DetectorX and the competing detectors at distances of (a) 15, (b) 30, and (c) 50 meters from the target area, respectively.

## 6. Conclusion

This study presents a new framework, DetectorX, for structural damage detection and related structural health monitoring applications. DetectorX introduces two novel modules: (1) a dynamic visual modality via a stem block and (2) a spiral pooling technique. The stem block induces two separate visual modalities based on the outputs of two DCNN models within DetectorX, which operates on the proposed event-based reward reinforcement learning, which constrains child and parent action as a single action leading to a reward. The dynamic visual modalities introduced in addition to the RGB input data, three distinct visual modalities, allow DetectorX to improve its perception, hence improving its overall resilience and flexibility in diverse environmental conditions. The framework's robustness is further augmented by incorporating spiral pooling, an online technique for augmenting images. The technique of spiral pooling leverages the intrinsic arrangement of spiral patterns to facilitate the introduction of varied feature maps in the training process. Utilizing this technique increases the variety of the feature representations and spatial linkages, contributing to the overall satisfactory performance of DetectorX in three experimental settings: (1) comparative, (2) robustness, and (3) field experiments, respectively.

Utilizing the PEER Hub Image-Net data, specifically, task 8, damage type in the comparative and robustness experiment, DetectorX outperformed detectors including YOLOX-m, EfficientDet-D5, YOLOv5-m, SSD500, and Faster-R-CNN, as well as two variants of DetectorX; thus, DetectorX-SB, and DetectorX-SP. DetectorX attained satisfactory results under varying evaluation metrics, including precision, recall, F1 score, average precision (AP), mean average precision (mAP), mean average recall (mAR), and competitive inference time and frame processing speed. Specifically, DetectorX attained a precision of 0.88, a recall of 0.84, and mAP and mAR of 0.76 and 0.73 compared to a competitive detector YOLOX-m, which had a precision and recall of 0.86 and 0.82, as well as mAP and mAR of 0.73 and 0.71, respectively. In the robustness test, DetectorX was favorable and demonstrated robustness in challenging conditions, including blurring, illumination, and noise conditions, respectively, compared to the other detectors. Even in extreme intensities of the robustness test in blurring, illumination, and noisy settings, DetectorX attained precisions and recalls of 0.74 and 0.71, 0.70 and 0.68, 0.72 and 0.70, respectively, than YOLOX-m, which was 0.71 and 0.69, 0.68 and 0.66, and 0.70 and 0.69, respectively. Further, in the field experiment with a micro drone, under two conditions: (1) variant micro drone speed to detection performance, and (2) three distinct distances to target area detection performance, DetectorX outperformed the other detection models with precision between 0.74 to 0.84, and recall between 0.70 to 0.83.

Overall, the findings of this comprehensive study highlight DetectorX as a promising, robust, and resilient detection framework across multiple experimental settings. DetectorX's superior performance in structural damage detection, robustness, micro drone speed-to-detection evaluation, and distance detection sets it apart from the competing detection models. However, DetectorX could benefit from model optimization techniques to increase the inference time and frame process speed. Also, the spiral pooling mechanisms increase the training time since the spiral operation occurs during training. A further optimized spiralling mechanism may be needed to reduce the training time.

## Declaration of competing interest

The authors declare that they have no known competing financial interests or personal relationships that could have appeared to influence the work reported in this paper.

## Data availability

The data used has been referenced in the paper, specifically reference [48], PEER Hub ImageNet: A Large-Scale Multiattribute Benchmark Data Set of Structural Images.

## Acknowledgments

This work is partly supported by the Department of Science and Technology of Sichuan Province, China, under Grant 2022NSFSC0509 and Research Funds for the Central Universities, China, under Grant ZYGX2021YGLH020.

## References

[1] W. Wang, C. Su, G. Han, H. Zhang, A lightweight crack segmentation network based on knowledge distillation, Journal of Building Engineering 76 (June) (2023), p. 107200, https://doi.org/10.1016/j.jobe.2023.107200.

[2] Y. Gao, J. Yang, H. Qian, K.M. Mosalam, Multiattribute multitask transformer framework for vision-based structural health monitoring, Computer-Aided Civil and Infrastructure Engineering (2023), pp. 1–20, https://doi.org/10.1111/mice.13067.

[3] E. Asadi Shamsabadi, C. Xu, A.S. Rao, T. Nguyen, T. Ngo, D. Dias-da-Costa, Vision transformer-based autonomous crack detection on asphalt and concrete surfaces, Automation in Construction 140 (December 2021) (2022), p. 104316, https://doi.org/10.1016/j.autcon.2022.104316.

[4] L. Yang, H. Huang, S. Kong, Y. Liu, A deep segmentation network for crack detection with progressive and hierarchical context fusion, Journal of Building Engineering 75 (May) (2023), p. 106886, https://doi.org/10.1016/j.jobe.2023.106886.

[5] M. Bolognini, G. Izzo, D. Marchisotti, L. Fagiano, M.P. Limongelli, E. Zappa, Vision-based modal analysis of built environment structures with multiple drones, Automation in Construction 143 (August) (2022), p. 104550, https://doi.org/10.1016/j.autcon.2022.104550.

[6] A.M. Roy, J. Bhaduri, DenseSPH-YOLOv5: An automated damage detection model based on DenseNet and Swin-Transformer prediction head-enabled YOLOv5 with attention mechanism, Advanced Engineering Informatics 56 (October 2022) (2023),, https://doi.org/10.1016/j.aei.2023.102007.

[7] M.A. Jayaram, Computer vision applications in construction material and structural health monitoring: A scoping review, Materials Today: Proceedings (xxxx) (2023),, https://doi.org/10.1016/j.matpr.2023.06.031.

[8] A. Soleymani, H. Jahangir, M.L. Nehdi, Damage detection and monitoring in heritage masonry structures: Systematic review, Construction and Building Materials 397 (January) (2023), p. 132402, https://doi.org/10.1016/j.conbuildmat.2023.132402.

[9] A.H. El Hakea, M.W. Fakhr, Recent computer vision applications for pavement distress and condition assessment, Automation in Construction 146 (November 2022) (2023), p. 104664, https://doi.org/10.1016/j.autcon.2022.104664.

[10] C.Z. Dong, F.N. Catbas, A review of computer vision–based structural health monitoring at local and global levels, Structural Health Monitoring 20 (2) (2021), pp. 692–743, https://doi.org/10.1177/1475921720935585.

[11] I.O. Agyemang, X. Zhang, I. Adjei-Mensah, B.L.Y. Agbley, L.D. Fiasam, B.C. Mawuli, C. Sey, Accelerating Classification on Resource-Constrained Edge Nodes Towards Automated Structural Health Monitoring, , in 2021 18th International Computer Conference on Wavelet Active Media Technology and Information Processing, ICCWAMTIP 2021, 2021, pp. 157–160. https://doi.org/10.1109/ICCWAMTIP53232.2021.9674071.

[12] N. Kheradmandi, V. Mehranfar, A critical review and comparative study on image segmentation-based techniques for pavement crack detection, Construction and Building Materials 321 (December 2021) (2022), p. 126162, https://doi.org/10.1016/j.conbuildmat.2021.126162.

[13] J. Deng, A. Singh, Y. Zhou, Y. Lu, V.C.S. Lee, Review on computer vision-based crack detection and quantification methodologies for civil structures, Construction and Building Materials 356 (September) (2022),, https://doi.org/10.1016/j.conbuildmat.2022.129238.

[14] N. Nagesh, K. Raisi, N.A. Valente, J. Benoit, T. Yu, A. Sabato, Deep learning augmented infrared thermography for unmanned aerial vehicles structural health monitoring of roadways, Automation in Construction 148 (October 2022) (2023), p. 104784, https://doi.org/10.1016/j.autcon.2023.104784.

[15] X.W. Ye, T. Jin, P.Y. Chen, Structural crack detection using deep learning–based fully convolutional networks, Advances in Structural Engineering 22 (16) (2019), pp. 3412–3419, https://doi.org/10.1177/1369433219836292.

[16] B.G. Pantoja-rosero, R. Achanta, K. Beyer, Damage-augmented digital twins towards the automated inspection of buildings, Automation in Construction 150 (March) (2023), p. 104842, https://doi.org/10.1016/j.autcon.2023.104842.

[17] A.A. Baffour, Z. Qin, Y. Wang, Z. Qin, K.K.R. Choo, Spatial self-attention network with self-attention distillation for fine-grained image recognition, Journal of Visual Communication and Image Representation 81 (2021), p. 103368, https://doi.org/10.1016/j.jvcir.2021.103368.

[18] Y. Zhou, A. Ji, L. Zhang, X. Xue, Attention-enhanced sampling point cloud network (ASPCNet) for efficient 3D tunnel semantic segmentation, Automation in Construction 146 (January 2022) (2023), p. 104667, https://doi.org/10.1016/j.autcon.2022.104667.

[19] I.O. Agyemang, X. Zhang, I. Adjei-mensah, J.R. Arhin, E. Agyei, Lightweight Real-time Detection of Components via a Micro Aerial Vehicle with Domain Randomization Towards Structural Health Monitoring, Periodica Polytechnica Civil Engineering 66 (2) (2022), pp. 1–16, [Online]. Available: https://doi.org/10.3311/PPci.18689

[20] A. Ramezani Dooraki, D.J. Lee, An innovative bio-inspired flight controller for quad-rotor drones: Quad-rotor drone learning to fly using reinforcement learning, Robotics and Autonomous Systems 135 (2021), p. 103671, https://doi.org/10.1016/j.robot.2020.103671.

[21] D. Falanga, K. Kleber, D. Scaramuzza, Dynamic obstacle avoidance for quadrotors with event cameras, Science Robotics 5 (40) (2020),, https://doi.org/10.1126/scirobotics.aaz9712.

[22] C. Gao, X. Wang, R. Wang, Z. Zhao, Y. Zhai, X. Chen, B.M. Chen, A UAV-based explore-then-exploit system for autonomous indoor facility inspection and scene reconstruction, Automation in Construction 148 (December 2022) (2023), p. 104753, https://doi.org/10.1016/j.autcon.2023.104753.

[23] I.O. Agyemang, X. Zhang, D. Acheampong, I. Adjei-Mensah, E. Opanin Gyamfi, W. Ayivi, C. Ijeoma Amuche, E.O. Gyamfi, J.R. Arhin, W. Ayivi, C.I. Amuche, Rpnet: Rotational Pooling Net for Efficient Micro Aerial Vehicle Trail Navigation, Engineering Applications of Artificial Intelligence 116 (May) (2022), p. 105468, https://doi.org/10.1016/j.engappai.2022.105468.

[24] Y. Yuan, J. Zhang, Q. Wang, Deep Gabor convolution network for person re-identification, Neurocomputing 378 (2020), pp. 387–398, https://doi.org/10.1016/j.neucom.2019.10.083.

[25] A. Loquercio, E. Kaufmann, R. Ranftl, A. Dosovitskiy, V. Koltun, D. Scaramuzza, Deep Drone Racing: From Simulation to Reality with Domain Randomization, IEEE Transactions on Robotics 36 (1) (2020), pp. 1–14, https://doi.org/10.1109/TRO.2019.2942989.

[26] I.O. Agyemang, X. Zhang, I. Adjei-Mensah, B.L.Y. Agbley, B.C. Mawuli, L.D. Fiasam, C. Sey, Accelerating trail navigation for unmanned aerial vehicle: A denoising deep-net with 3D-NLGL, Journal of Intelligent and Fuzzy Systems 43 (6) (2022), pp. 7277–7295, https://doi.org/10.3233/JIFS-220693.

[27] Q. Qiu, D. Lau, Real-time detection of cracks in tiled sidewalks using YOLO-based method applied to unmanned aerial vehicle ( UAV ) images, Automation in Construction 147 (May 2022) (2023), p. 104745, https://doi.org/10.1016/j.autcon.2023.104745.

[28] D.H. Kang, Y.J. Cha, Efficient attention-based deep encoder and decoder for automatic crack segmentation, Structural Health Monitoring (2021),, https://doi.org/10.1177/14759217211053776.

[29] J. Deng, A. Singh, Y. Zhou, Y. Lu, V.C.S. Lee, Review on computer vision-based crack detection and quantification methodologies for civil structures, Construction and Building Materials 356 (June) (2022),, https://doi.org/10.1016/j.conbuildmat.2022.129238.

[30] F. Alsakka, S. Assaf, I. El-Chami, M. Al-Hussein, Computer vision applications in offsite construction, Automation in Construction 154 (May) (2023), p. 104980, https://doi.org/10.1016/j.autcon.2023.104980.

[31] D. Ai, G. Jiang, S.K. Lam, P. He, C. Li, Computer vision framework for crack detection of civil infrastructure—A review, Engineering Applications of Artificial Intelligence 117 (August 2022) (2023), p. 105478, https://doi.org/10.1016/j.engappai.2022.105478.

[32] I.O. Agyemang, X. Zhang, I.A. Mensah, B.C. Mawuli, B.L.Y. Agbley, J.R. Arhin, Enhanced deep convolutional neural network for building component detection towards structural health monitoring, , in 2021 4th International Conference on Pattern Recognition and Artificial Intelligence, PRAI 2021, 2021, pp. 202–206. https://doi.org/10.1109/PRAI53619.2021.9551102.

[33] I.O. Agyemang, X. Zhang, I. Adjei-mensah, D. Acheampong, L. Delali, C. Sey, S. Banaamwini, D. Effah, Automated vision-based structural health inspection and assessment for post-construction civil infrastructure, Automation in Construction 156 (October) (2023), p. 105153, https://doi.org/10.1016/j.autcon.2023.105153.

[34] G. Ye, J. Qu, J. Tao, W. Dai, Y. Mao, Q. Jin, Autonomous surface crack identification of concrete structures based on the YOLOv7 algorithm, Journal of Building Engineering 73 (April) (2023), p. 106688, https://doi.org/10.1016/j.jobe.2023.106688.

[35] M. Dang, H. Wang, T.-H. Nguyen, L. Tightiz, L. Dinh Tien, T.N. Nguyen, N.P. Nguyen, CDD-TR: Automated concrete defect investigation using an improved deformable transformers, Journal of Building Engineering 75 (May) (2023), p. 106976, https://doi.org/10.1016/j.jobe.2023.106976.

[36] S. Katsigiannis, S. Seyedzadeh, A. Agapiou, N. Ramzan, Deep learning for crack detection on masonry façades using limited data and transfer learning, Journal of Building Engineering 76 (June) (2023), p. 107105, https://doi.org/10.1016/j.jobe.2023.107105.

[37] I.O. Agyemang, X. Zhang, I. Adjei-Mensah, B.C. Mawuli, B.L.Y. Agbley, L.D. Fiasam, C. Sey, On Salient Concrete Crack Detection Via Improved Yolov5, , in 2021 18th International Computer Conference on Wavelet Active Media Technology and Information Processing, ICCWAMTIP 2021, 2021, pp. 175–178. https://doi.org/10.1109/ICCWAMTIP53232.2021.9674153.

[38] A.D. Andrushia, N. Anand, T.M. Neebha, M.Z. Naser, E. Lubloy, Autonomous detection of concrete damage under fire conditions, , Automation in Construction, 140. 2022. https://doi.org/10.1016/j.autcon.2022.104364.

[39] X. Weng, Y. Huang, Y. Li, H. Yang, S. Yu, Unsupervised domain adaptation for crack detection, Automation in Construction 153 (April 2022) (2023), p. 104939, https://doi.org/10.1016/j.autcon.2023.104939.

[40] R. Li, Y. Yuan, W. Zhang, Y. Yuan, Unified Vision-Based Methodology for Simultaneous Concrete Defect Detection and Geolocalization, Computer-Aided Civil and Infrastructure Engineering 33 (7) (2018), pp. 527–544, https://doi.org/10.1111/mice.12351.

[41] E. Garilli, N. Bruno, F. Autelitano, R. Roncella, F. Giuliani, Automatic detection of stone pavement’s pattern based on UAV photogrammetry, Automation in Construction 122 (October 2020) (2021), p. 103477, https://doi.org/10.1016/j.autcon.2020.103477.

[42] W. Ding, H. Yang, K. Yu, J. Shu, Crack detection and quantification for concrete structures using UAV and transformer, Automation in Construction 152 (April) (2023),, https://doi.org/10.1016/j.autcon.2023.104929.

[43] C. Cheng, Z. Shang, Z. Shen, Automatic delamination segmentation for bridge deck based on encoder-decoder deep learning through UAV-based thermography, NDT and E International 116 (August) (2020), p. 102341, https://doi.org/10.1016/j.ndteint.2020.102341.

[44] S. Jiang, J. Zhang, Real-time crack assessment using deep neural networks with wall-climbing unmanned aerial system, Computer-Aided Civil and Infrastructure Engineering 35 (6) (2020), pp. 549–564, https://doi.org/10.1111/mice.12519.

[45] S. Tavasoli, X. Pan, T.Y. Yang, Real-time autonomous indoor navigation and vision-based damage assessment of reinforced concrete structures using low-cost nano aerial vehicles, Journal of Building Engineering 68 (January) (2023), p. 106193, https://doi.org/10.1016/j.jobe.2023.106193.

[46] I.O. Agyemang, X. Zhang, D. Acheampong, I. Adjei-Mensah, G.A. Kusi, B.C. Mawuli, B.L.Y. Agbley, Autonomous health assessment of civil infrastructure using deep learning and smart devices, Automation in Construction 141 (February) (2022), p. 104396, https://doi.org/10.1016/j.autcon.2022.104396.

[47] M. Tan, Q. V. Le, EfficientNet: Rethinking model scaling for convolutional neural networks, , in International Conference on Machine Learning, ICML 2019, 2019, 2019-June, pp. 10691–10700. [Online]. Available: https://proceedings.mlr.press/v97/tan19a/tan19a.pdf

[48] M. Tan, R. Pang, Q. V. Le, EfficientDet: Scalable and efficient object detection, , in Proceedings of the IEEE Computer Society Conference on Computer Vision and Pattern Recognition, 2020, pp. 10778–10787. https://doi.org/10.1109/CVPR42600.2020.01079.

[49] Y. Gao, K.M. Mosalam, PEER Hub ImageNet: A Large-Scale Multiattribute Benchmark Data Set of Structural Images, Journal of Structural Engineering 146 (10) (2020), p. 04020198, https://doi.org/10.1061/(asce)st.1943-541x.0002745.

[50] Z. Ge, S. Liu, F. Wang, Z. Li, J. Sun, YOLOX: Exceeding YOLO Series in 2021, , (2021),. [Online]. Available: http://arxiv.org/abs/2107.08430
[51] Ultralytics, YOLOv5: A state-of-the-art real-time object detection system, . https://doi.org/https://docs.ultralytics.com/yolov5/.
[52] W. Liu, D. Anguelov, D. Erhan, C. Szegedy, S. Reed, C.-Y. Fu, A.C. Berg, SSD: Single Shot MultiBox Detector, , in European Conference on Computer Vision, 2016, 1, pp. 398–413. https://doi.org/10.1007/978-3-319-46448-0.
[53] S. Ren, K. He, R. Girshick, J. Sun, Faster R-CNN: Towards Real-Time Object Detection with Region Proposal Networks, IEEE Transactions on Pattern Analysis and Machine Intelligence 39 (6) (2017), pp. 1137–1149, https://doi.org/10.1109/TPAMI.2016.2577031.